\documentclass[hidelinks,onefignum,onetabnum]{siamart220329}

\usepackage{amsfonts,amsmath,amssymb,amsopn}
\usepackage{xparse}[2019/10/11]
\usepackage{xcolor,cleveref}
\usepackage{graphicx}
\usepackage{epstopdf}

\usepackage{algorithm,algorithmicx,algpseudocode}
\usepackage{nicefrac,multirow}
\algrenewcommand\algorithmicrequire{\textbf{Input:}}
\algrenewcommand\algorithmicensure{\textbf{Output:}}

\Crefname{ALC@unique}{Line}{Lines}

\definecolor{darkgreen}{RGB}{0,128,0}
\definecolor{darkblue}{RGB}{0,0,128}
\definecolor{darkred}{RGB}{128,0,0}

\makeatletter
\newcommand{\thickhline}{%
    \noalign {\ifnum 0=`}\fi \hrule height 1pt
    \futurelet \reserved@a \@xhline
}
\makeatother

\DeclareDocumentCommand \prArg{mm}
{(
\IfNoValueTF{#2}{#1}{#1 \mid #2}
)}

\DeclareDocumentCommand \newProbabilityFormat{r<>m}
{
	\DeclareDocumentCommand #1 {e{_}e{^}>{\SplitArgument{1}{|}}d()}
	{
		\IfNoValueTF{##1}
		{
			\IfNoValueTF{##2}
			{\IfNoValueTF{##3}{#2}{#2\prArg##3}}
			{\IfNoValueTF{##3}{#2^{##2}}{#2^{##2}\prArg##3}}
		}
		{
			\IfNoValueTF{##2}
			{\IfNoValueTF{##3}{#2_{##1}}{#2_{##1}\prArg##3}}
			{\IfNoValueTF{##3}{#2_{##1}^{##2}}{#2_{##1}^{##2}\prArg##3}}
		}
	}
}

\DeclareDocumentCommand \fVector {m} {\boldsymbol{#1}}
\DeclareDocumentCommand \fMatrix {m} {\boldsymbol{#1}}

\DeclareDocumentCommand \fFunction {m} {{#1}}
\DeclareDocumentCommand \fSet {m} {\mathcal{#1}}

\DeclareDocumentCommand \fLimOperator {m} {\mathop{\vphantom{\lim}\mathchoice {\hbox{#1}} {\vcenter{\hbox{#1}}}{#1}{#1}}\displaylimits}

\DeclareDocumentCommand \newScalar{r<>m}
{
	\DeclareDocumentCommand #1 {} {{#2}}
}

\DeclareDocumentCommand \newVector{r<>m}
{
	\DeclareDocumentCommand #1 {} {\fVector{#2}}
}

\DeclareDocumentCommand \newMatrix{r<>m}
{
	\DeclareDocumentCommand #1 {} {\fMatrix{#2}}
}

\DeclareDocumentCommand \newProbability{r<>m}
{
	\newProbabilityFormat<#1>{#2}
}

\DeclareDocumentCommand \newFunction{r<>m}
{
	\DeclareDocumentCommand #1 {} {\fFunction{#2}}
}

\DeclareDocumentCommand \newSet{r<>m}
{
	\DeclareDocumentCommand #1 {} {\fSet{#2}}
}

\makeatletter
\def\cleartheorem#1{%
    \expandafter\let\csname#1\endcsname\relax
    \expandafter\let\csname c@#1\endcsname\relax
}
\makeatother

\DeclareDocumentCommand \argmin {} {\fLimOperator{argmin}}

\DeclareMathOperator{\expect}{\mathbb{E}}

\DeclareMathOperator{\diag}{diag}

\DeclareDocumentCommand \reals {} {\mathbb{R}}

\newScalar<\eps>{\varepsilon}

\newVector<\vA>{a}
\newVector<\vB>{b}
\newVector<\vC>{c}
\newVector<\vD>{d}
\newVector<\vE>{e}
\newVector<\vF>{f}
\newVector<\vG>{g}
\newVector<\vH>{h}
\newVector<\vI>{i}
\newVector<\vJ>{j}
\newVector<\vK>{k}
\newVector<\vL>{\ell}
\newVector<\vM>{m}
\newVector<\vN>{n}
\newVector<\vP>{p}
\newVector<\vQ>{q}
\newVector<\vR>{r}
\newVector<\vS>{s}
\newVector<\vT>{t}
\newVector<\vU>{u}
\newVector<\vV>{v}
\newVector<\vW>{w}
\newVector<\vX>{x}
\newVector<\vY>{y}
\newVector<\vZ>{z}
\newVector<\vAlpha>{\alpha}
\newVector<\vBeta>{\beta}
\newVector<\vGamma>{\gamma}
\newVector<\vDelta>{\delta}
\newVector<\vEpsilon>{\varepsilon}
\newVector<\vEps>{\varepsilon}
\newVector<\vZeta>{\zeta}
\newVector<\vEta>{\eta}
\newVector<\vTheta>{\theta}
\newVector<\vTh>{\theta}
\newVector<\vKappa>{\kappa}
\newVector<\vLambda>{\lambda}
\newVector<\vMu>{\mu}
\newVector<\vNu>{\nu}
\newVector<\vXi>{\xi}
\newVector<\vPi>{\pi}
\newVector<\vRho>{\rho}
\newVector<\vSigma>{\sigma}
\newVector<\vTau>{\tau}
\newVector<\vUpsilon>{\upsilon}
\newVector<\vPhi>{\varphi}
\newVector<\vChi>{\chi}
\newVector<\vPsi>{\psi}
\newVector<\vOmega>{\omega}

\newMatrix<\mA>{A}
\newMatrix<\mB>{B}
\newMatrix<\mC>{C}
\newMatrix<\mD>{D}
\newMatrix<\mE>{E}
\newMatrix<\mF>{F}
\newMatrix<\mG>{G}
\newMatrix<\mH>{H}
\newMatrix<\mI>{I}
\newMatrix<\mJ>{J}
\newMatrix<\mK>{K}
\newMatrix<\mL>{L}
\newMatrix<\mM>{M}
\newMatrix<\mN>{N}
\newMatrix<\mP>{P}
\newMatrix<\mQ>{Q}
\newMatrix<\mR>{R}
\newMatrix<\mS>{S}
\newMatrix<\mT>{T}
\newMatrix<\mU>{U}
\newMatrix<\mV>{V}
\newMatrix<\mW>{W}
\newMatrix<\mX>{X}
\newMatrix<\mY>{Y}
\newMatrix<\mZ>{Z}
\newMatrix<\mGamma>{\Gamma}
\newMatrix<\mDelta>{\Delta}
\newMatrix<\mTheta>{\Theta}
\newMatrix<\mLambda>{\Lambda}
\newMatrix<\mXi>{\Xi}
\newMatrix<\mPi>{\Pi}
\newMatrix<\mSigma>{\Sigma}
\newMatrix<\mUpsilon>{\Upsilon}
\newMatrix<\mPhi>{\Phi}
\newMatrix<\mPsi>{\Psi}
\newMatrix<\mOmega>{\Omega}

\newSet<\sA>{A}
\newSet<\sB>{B}
\newSet<\sC>{C}
\newSet<\sD>{D}
\newSet<\sE>{E}
\newSet<\sF>{F}
\newSet<\sG>{G}
\newSet<\sH>{H}
\newSet<\sI>{I}
\newSet<\sJ>{J}
\newSet<\sK>{K}
\newSet<\sL>{L}
\newSet<\sM>{M}
\newSet<\sN>{N}
\newSet<\sP>{P}
\newSet<\sQ>{Q}
\newSet<\sR>{R}
\newSet<\sS>{S}
\newSet<\sT>{T}
\newSet<\sU>{U}
\newSet<\sV>{V}
\newSet<\sW>{W}
\newSet<\sX>{X}
\newSet<\sY>{Y}
\newSet<\sZ>{Z}
\DeclareMathOperator{\sSpan}{span}

\newsiamremark{remark}{Remark}
\newsiamremark{hypothesis}{Hypothesis}
\crefname{hypothesis}{Hypothesis}{Hypotheses}
\newsiamthm{claim}{Claim}

\headers{Projective Integral Updates for Variational Inference}{J. A. Duersch}
\title{Projective Integral Updates for High-Dimensional Variational Inference}
\ifpdf
  \hypersetup{
    pdftitle={Projective Integral Updates for High-Dimensional Variational Inference},
    pdfauthor={J. A. Duersch}
  }
  \DeclareGraphicsExtensions{.eps,.pdf,.png,.jpg}
\else
  \DeclareGraphicsExtensions{.eps}
\fi

\author{Jed A. Duersch\thanks{Sandia National Laboratories, Livermore, CA (\email{jaduers@sandia.gov}).}}

\newScalar<\nPa>{d}
\newScalar<\nDim>{d}
\newScalar<\iPa>{i}
\newScalar<\iDim>{i}
\newScalar<\nBit>{n_b}
\newScalar<\iBit>{j}
\newScalar<\iIter>{t}

\newScalar<\iQuad>{q}
\newScalar<\iSample>{q}
\newScalar<\nQuad>{n_q}
\newScalar<\nSample>{n_s}

\newScalar<\iEpoch>{e}
\newScalar<\nEpoch>{n_e}

\newScalar<\iBasis>{j}
\newScalar<\nBasis>{n_{j}}
\newScalar<\nExact>{n_{\text{exact}}}

\newScalar<\iBlock>{b}
\newScalar<\nBlock>{n_b}

\newScalar<\iData>{c}
\newScalar<\nData>{n_c}
\newScalar<\nTrain>{\nData}

\newScalar<\lenBlock>{\ell}

\newScalar<\sAnneal>{\gamma}
\newScalar<\hPr>{h_p}
\newScalar<\sigmaPr>{\sigma_p}
\newScalar<\varPr>{\sigma_p^2}
\newScalar<\logitPr>{\zeta_p}

\newMatrix<\sTh>{\Theta}
\newMatrix<\Th>{\Theta}

\newScalar<\logitSrcZero>{\vZeta^{(0)}}
\newScalar<\logitSrcOne>{\vZeta^{(1)}}

\newScalar<\logitTgtZero>{\hat{\vZeta}^{(0)}}
\newScalar<\logitTgtOne>{\hat{\vZeta}^{(1)}}

\newScalar<\fZero>{s^{(0)}}
\newScalar<\fOne>{s^{(1)}}

\newScalar<\fZeroTarget>{s_{\text{tgt}}^{(0)}}
\newScalar<\fOneTarget>{s_{\text{tgt}}^{(1)}}

\newScalar<\pNZZero>{p_{\text{NZ}}^{(0)}}
\newScalar<\pNZOne>{p_{\text{NZ}}^{(1)}}

\newScalar<\aInit>{\alpha_0}
\newScalar<\hInit>{h_0}
\newScalar<\aMax>{\alpha_\text{max}}
\newScalar<\tauMax>{\tau_\text{max}}

\newScalar<\iMode>{m}

\newVector<\vPa>{\theta}
\newVector<\vGh>{\hat{g}}
\newVector<\vHh>{\hat{h}}

\newMatrix<\mFactor>{A}
\newSet<\tC>{\mathcal{C}}

\newSet<\sData>{\mathcal{D}}
\newSet<\sTrain>{\mathcal{T}}
\newSet<\sValid>{\mathcal{V}}
\newSet<\sFeasible>{\mathcal{F}}

\newProbability<\pP>{\mathbf{p}}
\newProbability<\pQ>{\mathbf{q}}
\newProbability<\pR>{\mathbf{r}}
\newProbability<\pN>{\mathcal{N}}
\newProbability<\pL>{\mathbf{Laplace}}
\newProbability<\pD>{\mathbf{\delta}}

\newFunction<\fLoss>{\mathcal{L}}
\newScalar<\zeroLoss>{\ell}
\newFunction<\fInt>{f}
\newFunction<\fUni>{\varphi}
\newFunction<\fBasis>{f}
\newFunction<\fResidual>{r}
\newFunction<\bit>{\mathbf{bit}}
\newFunction<\bits>{\mathbf{bits}}
\newFunction<\sgn>{\text{sign}}
\newFunction<\logit>{\text{logit}}
\DeclareDocumentCommand \xor {} {\fLimOperator{\ensuremath{\mathbf{xor}}}}
\DeclareDocumentCommand \and {} {\fLimOperator{\ensuremath{\land}}}

\newVector<\trainSymbol>{\mathcal{T}}

\newFunction<\fSimplexQuad>{\mathbf{simplex\_quadrature}}
\newFunction<\fCrossPolyQuad>{\mathbf{quadrature\_sequence}}
\newFunction<\fQuadraticApprox>{\mathbf{quadratic\_approx}}
\newFunction<\fStep>{\mathbf{step}}
\newFunction<\fTrain>{\mathbf{train}}
\newFunction<\fEpoch>{\mathbf{variational\_epoch}}
\newFunction<\fVariationalUpdate>{\mathbf{variational\_update}}

\DeclareDocumentCommand \KL{mm} {\mathbb{D}\!\left[\,#1\,\middle\|\,#2\,\right]}

\DeclareDocumentCommand \dsum {} {\fLimOperator{\ensuremath{\oplus}}}

\DeclareDocumentCommand \tElse{} {\text{else}}
\DeclareDocumentCommand \tConst{} {\text{\small{const}}}

\DeclareDocumentCommand\precaption{}{\vspace{-3mm}}
\DeclareDocumentCommand\postcaption{}{\vspace{-3mm}}

\begin{document}

\maketitle


\begin{abstract}
Variational inference is an approximation framework for Bayesian inference that seeks to improve quantified uncertainty in predictions by optimizing a simplified distribution over parameters to stand in for the full posterior.
Capturing model variations that remain consistent with training data enables more robust predictions by reducing parameter sensitivity.
This work introduces a fixed-point optimization for variational inference that is applicable when every feasible log density can be expressed as a linear combination of functions from a given basis.
In such cases, the optimizer becomes a fixed-point of projective integral updates.
When the basis spans univariate quadratics in each parameter, the feasible distributions are Gaussian mean-fields and the projective integral updates yield quasi-Newton variational Bayes (QNVB).
Other bases and updates are also possible.
Since these updates require high-dimensional integration,
this work begins by proposing an efficient quasirandom sequence of quadratures for mean-field distributions.
Each iterate of the sequence contains two evaluation points that combine to correctly integrate all univariate quadratic functions and,
if the mean-field factors are symmetric, all univariate cubics.
More importantly, averaging results over short subsequences achieves periodic exactness on a much larger space of multivariate polynomials of quadratic total degree.
The corresponding variational updates require 4 loss evaluations with standard (not second-order) backpropagation to eliminate error terms from over half of all multivariate quadratic basis functions.
This integration technique is motivated by first proposing stochastic blocked mean-field quadratures, which may be useful in other contexts.
A PyTorch implementation of QNVB allows for better control over model uncertainty during training than competing methods.
Experiments demonstrate superior generalizability for multiple learning problems and architectures.
\end{abstract}

\begin{keywords}
variational inference,
Gaussian mean-field,
Hessian approximation,
quasi-Newton,
spike-and-slab,
quadrature,
cubature,
Hadamard basis
\end{keywords}

\begin{MSCcodes}
62F30 
65C05 
65D32 
65K10 
\end{MSCcodes}

\section{Introduction}
\label{sec:introduction}

Variational inference is an optimization-based approach to discover parameter domains that dominate the Bayesian posterior with origins in statistical physics \cite{Mezard1987,Parisi1988,Bishop2006,Hennig2011,Blei2017,Zhang2018}. 
One of the challenges of quantifying prediction uncertainty for high-dimensional models is how to reliably characterize model uncertainty as it evolves during training.
For high-dimensional model classes, mean-field distributions provide a simple and scalable method for tracking a component of model uncertainty
and thereby capturing a useful contribution to uncertainty in predictions at a reduced computational cost \cite{Anderson1987,Opper1996}.
In principle, optimizing the variational objective requires repeatedly integrating the log-likelihood of the training data as the variational density changes.
Therefore, this investigation began in pursuit of an efficient quadrature framework capable of capturing the primary contributions to the shape of mean-field densities from only a handful of function evaluations.
To incorporate these integral approximations into more efficient variational updates, this work goes on to develop a fixed-point formulation of the variational objective using projective integral updates,
which use the current variational density during optimization to compute loss expectations against certain basis functions.
These expectations then inform the structure of the next variational density.

The aim of this work is to improve the efficiency and generalizability of variational inference for high-dimensional model classes,
and the fixed-point formulation of the variational objective provides an alternative analytic framework to calibrate and control model uncertainty during optimization.
This framework can be applied if every log density from the family of feasible variational distributions can be written as a linear combination of functions from a given basis.
For example, a basis for all univariate quadratic polynomials over the model parameters captures all Gaussian mean-field approximations
and the projective integral updates recover quasi-Newton steps.
However, since these updates act on averaged gradients and Hessian diagonals,
this work first offers two efficient numerical quadratures for high-dimensional mean-field distributions.

The first approach, stochastic blocked mean-field quadratures, may be useful for learning architectures that, based on the model's computational structure,
allow us to identify key blocks of parameters that may contain important correlations.
Blocked quadratures allow some correlations to be captured, while still remaining feasible for high-dimensional model classes.

The second approach, derived as a quasirandom modification of the first,
gains an additional property of periodic exactness on much larger bases.
To put this property in perspective, we might expect that a quadrature taking 4 function evaluations in $d$ parameter dimensions, comprising $4(d+1)$ degrees of freedom, should only integrate the same number of basis functions.
Yet, by averaging a consecutive pair of two-point quadratures from a quasirandom sequence,
the result exactly integrates $1 + 2d + \frac{1}{4}d^2$ quadratic total-degree basis functions, i.e.~more than half.
These quadrature sequences may support adaptive integration to suppress error as needed by adding evaluations from the quadrature sequence while keeping previous computations.

These quadrature formulas facilitate efficient approximations of the expected gradient and Hessian diagonal using 
related linear functionals that act on only a few gradient evaluations from standard backpropagation.

\subsection{Contributions}

The key contributions of this work are
1.~two numerical integration schemes that are suitable for mean-field and blocked mean-field densities,
2.~the derivation of projective integral updates from a fixed-point formulation of variational inference, and
3.~implementation details for quasi-Newton variational Bayes (QNVB)\footnote{A PyTorch version is available at \url{https://github.com/sandialabs/qnvb}.},
a training algorithm using projective integral updates for Gaussian mean-fields that is designed to overcome practical challenges for deep learning.

\paragraph{Efficient Numerical Integration}
Both of the proposed integration approaches proceed by partitioning parameters into small blocks.
The first approach simply requires equal-weight sigma-point quadratures \cite{Uhlmann1995,Menegaz2015} within each block.
Provided that all blocks use the same number of function evaluations, the evaluation coordinates can be permuted uniformly at random and concatenated.
Doing so retains the same exactness property within each block, but also yields an expectation matching the tensor product cubature over all blocks.
This allows efficient integration of blocked mean-field distributions, which could contain more comprehensive factors that track correlations within each block, rather than only using products of univariate densities.

The second approach exchanges pseudorandom concatenation for quasirandom sequences.
Each element of the sequence is a 2-point quadrature that exactly integrates all linear combinations of univariate quadratics.
When the factor distributions are symmetric, we obtain Gaussian quadratures that integrate all linear combinations of univariate cubics.
Averaging over a subsequence that contains an integer multiple of $2^b$ elements, where $b = 1, 2, \ldots, \lceil \log_2(\nPa) \rceil$,
yields exactness on an extra $\nPa^2 \frac{2^b - 1}{2^{b+1}}$ dimensions of the function space comprising quadratic total degree polynomials.
This effect is due to the quasirandom sequence creating a hierarchy of overlapping blocks that contain tensor product cubatures from the underlying quadratures.
By iterating through the quadrature sequence in concert with randomly permuted training data, we obtain a method that efficiently cancels persistent errors.

\paragraph{Projective Integral Updates}
Updates to the variational density can be formulated as projections of the log-posterior onto a compatible basis.
For Gaussian mean-field distributions, this analysis yields quasi-Newton updates to the mean of the variational density.
Since the projective integral updates are derived from the variational basis, the same analytic framework facilitates other updates for different variational bases.
For example, a closely-related derivation provides updates for Dirac-Gauss mixtures, i.e.~spike and slab distributions \cite{Polson2018,Bai2020,Jantre2021}, which associate parameter zeros with finite probabilities.

\paragraph{Quasi-Newton Variational Bayes}
In practice, Hessian diagonals are often small or negative, yielding large or non-descending updates.
Thus, to enforce descent steps, QNVB incorporates the Hessian averaging technique used in AdaHessian \cite{yao2021adahessian}.
Further, the adaptive learning rate scheme used by Adam \cite{Kingma2014} is adapted to enforce safe upper bounds on expected parameter perturbations
using the same hyperparameters ($\beta_1,\, \beta_2,\, \varepsilon$).
QNVB includes additional hyperparameters that control the scales of parameter uncertainty during training.

Experiments include: 1.~image classification using ResNet18 \cite{he2016deep} for CIFAR-10 \cite{Krizhevsky2009},
2.~natural language processing using Tensorized Transformer \cite{ma2019tensorized} with the Penn Treebank (PTB) dataset \cite{marcus1993building}, and
3.~recommendation learning using DLRM \cite{naumov2019deep} for the Criteo Ad Kaggle dataset.
QNVB demonstrates state-of-the-art competitiveness with alternative algorithms, including: stochastic gradient descent (SGD) \cite{Robbins1951}, Adam, AdaGrad \cite{duchi2011adaptive}, AdaHessian, and stochastic gradient variational Bayes (SGVB) \cite{kingma2013auto}.

\subsection{Organization}

\Cref{sec:background} provides background on variational inference, mean field distributions, and related approaches to numerical integration.
\Cref{sec:quadratures} proposes and analyzes stochastic blocked mean-field quadratures and quasirandom quadrature sequences for high-dimensional numerical integration.
\Cref{sec:variational} analyzes the fixed-point formulation that bridges efficient integration with Gaussian mean-field updates, Dirac-Gauss mixtures,
and finishes by discussing implementation details for QNVB.
\Cref{sec:experiments} provides numerical experiments and concluding remarks.

\section{Background}
\label{sec:background}

Bayesian inference provides an attractive paradigm to quantify prediction uncertainty by capturing a distribution over models that could explain the available data.
Given a training dataset $\sData$,
a model class with parameters $\vPa$,
prior belief $\pP(\vPa)$,
and likelihood $\pP(\sData \mid \vPa)$,
we obtain the posterior from Bayes' theorem:
\begin{align*}
\pP(\vPa \mid \sData) = \frac{\pP(\sData \mid \vPa)\pP(\vPa)}{\pP(\sData)}
\quad\text{where}\quad
\pP(\sData) = \int \pP(\sData \mid \vPa)\,d\pP(\vPa)
\end{align*}
is the model evidence.
Unfortunately, when the likelihood function is complicated, especially for high-dimensional architectures,
capturing the shape of the posterior becomes intractable due to limited computational resources.

Variational inference mitigates this issue by approximating the posterior with a simpler distribution, $\pQ(\vPa \mid \vPhi)$.
The variational parameters $\vPhi$ characterize the approximate shape of a posterior-dominant region of parameter space.
We discover these domains by optimizing a variational objective, such as the maximizing the evidence lower bound (ELBO).
Since the ELBO optimizer also minimizes the Kullback-Leibler (KL) divergence \cite{Kullback1951}: 
\begin{align}
\label{eq:objective}
\KL{\pQ(\vPa \mid \vPhi)}{\pP(\vPa \mid \sData)} = \int \log\!\left( \frac{\pQ(\vPa \mid \vPhi)}{\pP(\vPa \mid \sData)} \right)\,d\pQ(\vPa \mid \vPhi),
\end{align}
which is the excess information created by replacing the posterior distribution with a feasible approximation \cite{Duersch2020}.

Provided the dataset is composed of independent cases from the data-generating process, $\sData = \left\{ \vD_\iData \mid \iData \in [\nData]\right\}$,
the optimizer of \Cref{eq:objective} can be written as a sum of integrals over each case $\iData$:
\begin{align*}
\vPhi^* = \argmin_{\vPhi} \quad \KL{\pQ(\vPa \mid \vPhi)}{\pP(\vPa)} - \sum_{\iData=1}^\nData \int \log \pP(\vD_\iData \mid \vPa)\,d\pQ(\vPa \mid \vPhi).
\end{align*}
Notably, this construction does not depend on the model-evidence integral.
Optimization does, however, require repeatedly evaluating the log-likelihood integral as the variational distribution evolves.

This work targets mean-field variational densities because they offer the most scalable approach for high-dimensional model classes.
Given $\nPa$ parameters, indexed%
\footnote{Parameters are enumerated from zero to be consistent with the analysis in \Cref{sec:quadratures}.}
as $i \in \{0, 1, \ldots, \nPa-1\}$, mean-field distributions take the form
\begin{align}
\label{eq:mean-field}
\pQ(\vPa \mid \vPhi) = \prod_{i=0}^{\nPa-1} \pQ(\vPa_i \mid \vPhi_i),
\end{align}
where each $\vPhi_i$ may be a block containing several variational parameters that describe the shape of each density factor.
The simplicity of this structure enables the efficient high-dimensional numerical quadratures developed in \Cref{sub:quasirandom}.

\subsection{Monte Carlo Versus Quadratures}

There are generally two approaches to account for probability in numerical integration, stochastic methods and deterministic methods.
In the first case, randomized sampling matches integral contributions \textit{in expectation} over pseudorandom events that generate function evaluations.
Monte Carlo methods, including Markov chain Monte Carlo (MCMC) \cite{Neal1992,Gilks1995} as well as tempered variations that achieve better posterior convergence \cite{Catanach2020,Latz2021},
are entirely stochastic, mapping pseudorandom numbers to a set of samples from the posterior distribution.
As the sample size $\nSample$ increases, the scale of the integration error drops as $\varepsilon(\nSample) \approx \varepsilon(1)\nSample^{-\nicefrac{1}{2}}$.
The drawback of sampling is that, although it will eventually produce integral approximations with arbitrarily small error,
the number of function evaluations needed can be quite large.

Deterministic approaches are the domain of typical numerical quadrature formulas.
For integration in multiple dimensions, these are often called \textit{cubature} methods.
We solve for $\nQuad$ evaluation locations paired with weights, $\left\{ (\vPa^{(\iQuad)}, w_{\iQuad}) \mid \iQuad \in [\nQuad] \right\}$, that exactly integrate some basis of functions,
$\mPhi = \left\{ \fBasis_\iBasis(\cdot) \mid \iBasis \in [\nExact] \right\}$, so that 
\begin{align*}
Q\left[ f \right] = \sum_{\iQuad = 1}^{\nQuad} w_{\iQuad} f(\vPa^{(\iQuad)}) \approx \int f(\vPa)\,d\pQ(\vPa)
\quad\text{and}\quad
Q\left[ \fBasis_\iBasis \right] = \int \fBasis_\iBasis(\vPa)\,d\pQ(\vPa)
\end{align*}
becomes exact for all $\iBasis \in [\nExact]$.
For example, Uhlmann's sigma-points \cite{Uhlmann1995} include $2d+1$ evaluations consisting of a central node and cross-polytope vertices (as signed standard basis vectors).
By appropriately scaling and translating these nodes, it is possible to exactly integrate all quadratic total-degree polynomials against a distribution with a known mean and covariance.

Smolyak quadratures \cite{Smolyak1963,Gerstner1998,Petras2003} are efficient formulas to generate high-degree integration approximations in a few dimensions.
Going back to the work of McNamee and Stenger on symmetric cubature formulas \cite{mcnamee1967construction},
followed by Novak and Ritter \cite{Novak1999},
it is possible to construct high-order cubatures by applying all signed permutations to a generator set of evaluation nodes.
This results in $\nQuad \approx \frac{(2\nPa)^k}{k!}$ quadrature nodes that exactly integrate $2k+1$ total-degree polynomials in $\nPa$ dimensions
against a weight function with the structure of \Cref{eq:mean-field}.
This technique has been incorporated into practical higher-order integration for expectation propagation by Kokkala, Solin, and S\"{a}rkk\"{a} \cite{kokkala2015sigma},
an alternative Bayesian approximation framework that optimizes the reversed KL-divergence.
 
Unfortunately, even taking $k=1$ is too expensive for current learning architectures, where $\nPa$ often ranges from $10^5$ to $10^{10}$ or more.
Since optimizing the variational distribution requires evaluating as many integrals as there are training data, this approach is not feasible.
We desire an approach that 1.~only uses a few function evaluations per integral, 2.~exactly integrates basis functions that dominate the shape of the mean-field distribution,
and 3.~efficiently suppresses the unavoidable errors associated with basis functions that we cannot afford to integrate exactly.

\subsection{Quasi-Monte Carlo Integration}

Quasi-Monte Carlo methods find middle ground by incorporating both pseudorandom and deterministic aspects within the sample-generating process.
Caflisch \cite{Caflisch1998} provides an overview targeting the perspective of applied mathematicians and numerical analysts.
For example, we can use two evaluation points, $\vPa^{(1)} = \mu - \delta$ and $\vPa^{(2)} = \mu + \delta$, with equal weights, $w_1 = w_2 = \frac{1}{2}$, to exactly integrate all affine functions against a distribution with mean $\mu$.
This is what Caflisch calls an \textit{antithetic pair}, i.e.~a pair of evaluation points balanced about the mean of the distribution.
Other moment-matching methods extend this simple technique.
If we know the mean of the distribution to be integrated, then we can adjust a set of samples to exactly integrate all affine functions by applying a simple translation of sample coordinates.
Given a set of samples, $\mTheta = \left\{ \vPa^{(\iSample)} \sim \pQ(\vPa) \mid \iSample \in [\nSample] \right\}$, and a known mean, $\expect_{\vQ}[\vPa]=\vMu$,
we can translate the samples as
\begin{align*}
\vPa^{(\iSample)'} = \vPa^{(\iSample)} - \hat{\vMu} + \vMu 
\quad\text{where}\quad
\hat{\vMu} = \frac{1}{\nSample} \sum_{\iSample=1}^\nSample \vPa^{(\iSample)}.
\end{align*}
If we also have known diagonal covariance, e.g.~$\expect_{\vQ}[(\vPa^{(\iSample)}-\vMu)(\vPa^{(\iSample)}-\vMu)^T]=\diag(\vSigma^2)$,
then we can use an affine transformation to exactly integrate all linear combinations of univariate quadratic polynomials,
\begin{align*}
\vPa^{(\iSample)'}  = (\vPa^{(\iSample)} - \hat{\vMu}) \ast \hat{\vSigma}^{-1} \ast \vSigma + \vMu
\quad\text{where}\quad
\hat{\vSigma}^2 = \frac{1}{\nSample} \sum_{\iSample}^{\nSample} (\vPa^{(i)} - \hat{\vMu})^2.
\end{align*}
The operator $\ast$ represents the Hadamard (elementwise) product and exponents are also applied elementwise.
As the set of samples becomes large, the sample moments converge to the true moments, making the correction increasingly modest, so that the same convergence properties of Monte Carlo hold.
Unfortunately, this approach requires having the full set of sample locations before the correction can be made,
whereas the quasirandom quadrature sequences described in \Cref{sub:quasirandom} can be accumulated to match more basis functions as more function evaluations become available.

Beylkin \cite{Beylkin2005} also examines numerical algorithms in high dimensions
and Dick \cite{Dick2013} provides a recent overview of quasi-Monte Carlo methods for high-dimensional integration over the unit cube, $[0,1]^d$.
Trefethen \cite{Trefethen2017,Trefethen2022} examines high-dimensional integration methods that aim to subdue the curse of dimensionality,
observing that it is the special structure of certain problem-dependent integrands,
deviating from the anisotropy of the hypercube,
that allows some quadrature formulas to avoid the exponential increase in evaluation nodes needed for tensor-product cubatures.

\subsection{Latin Hypercube Sampling}
\begin{figure}[ht]
	\centering
	\includegraphics[width=0.50\textwidth]{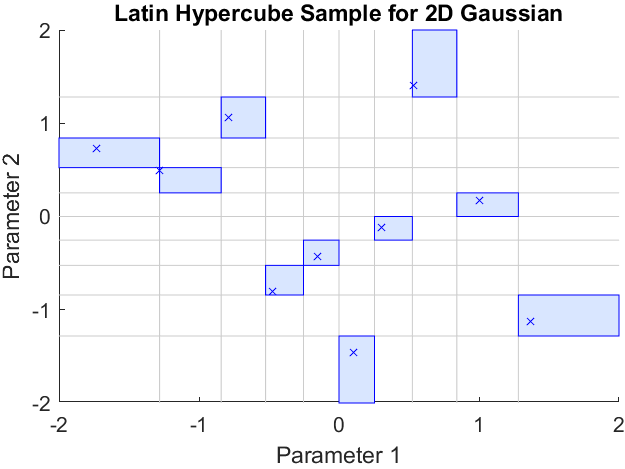}
	\precaption
	\caption{
	Latin hypercube illustration for 2D Gaussian separated into 10 equal-probability subintervals for each dimension.
	The quasirandom partition of samples produces a set of evaluation points that are more evenly distributed in each dimension.
	Samples are composed by randomly permuting the source partitions in each coordinate and concatenating the results.
	If the distribution is independent in each coordinate,
	the expectation matches the exact integral.
	}
	\postcaption
  \label{fig:latin}
\end{figure}

Latin hypercube sampling \cite{McKay2000} is another quasi-Monte Carlo approach that closely relates to this work.
It is a form of stratified sampling that matches, in expectation, stratified sampling on the Cartesian product of subsets in each dimension.
If we break a single dimension of the integral domain into subsets of equal probability,
and ensure that an equal number of samples are drawn from each subset,
then we obtain a more precise integral approximation in that dimension,
because the samples equally represent components that are analytically equivalent contributions.
The key insight of Latin hypercube sampling is that if we obtain such samples from each dimension independently within a mean-field distribution,
permuting samples within each dimension uniformly at random,
the expectation of the result still matches the product of integrals over all dimensions.
\Cref{fig:latin} provides an illustration of this technique in two dimensions.
The same idea drives the development and analysis of stochastic blocked mean-field quadratures in \Cref{sec:quadratures}.

\section{Analysis of Blocked Mean-Field Quadratures}
\label{sec:quadratures}

\Cref{sub:block_mean_field} begins with a discussion of blocked mean-field quadratures,
providing a stepping stone to the more powerful quasirandom sequences discussed in \Cref{sub:quasirandom}.
See \Cref{sec:numerical_integration} for numerical integration experiments for various mean-field distributions with corresponding basis functions.

\subsection{Stochastic Blocked Mean-Field Quadratures}
\label{sub:block_mean_field}

Just as a mean-field distribution is a product of univariate distributions in each parameter,
we can define a blocked mean-field distribution as a product of distributions over parameter blocks.
Formally, when we define such a blocking structure,
we decompose the parameter vector space into a direct sum of orthogonal subspaces,
\begin{align}
\label{eq:directsum}
\sTh = \dsum_{\iBlock=1}^\nBlock \sTh_\iBlock
\quad\text{so that}\quad
\vPa_\iBlock \in \sTh_\iBlock
\end{align}
indicates the parameters within a single block $\iBlock$, rather than an individual parameter.
With an appropriate basis, any parameter state can be represented by concatenating these components.
This facilitates a modest generalization of mean-field distributions to more comprehensive representations of parameter uncertainty within each block as
\begin{align}
\label{eq:blockmeanfield}
\pQ(\vTh) = \prod_{\iBlock=1}^\nBlock \pQ(\vTh_\iBlock)
\quad\text{where}\quad
\vPa = \left[ \vPa_1^T\,\vPa_2^T\,\cdots\,\vPa_\nBlock^T \right]^T \in \sTh.
\end{align}
By disregarding correlations between blocks, these densities remain scalable to high dimensions.
Just as Latin hypercube sampling concatenates stratified samples in each coordinate by using uniformly random permutations that match, in expectation, higher-dimensional stratification,
we can apply the same insight to equal-weight quadratures within each block to obtain composite quadratures with expectations that match the tensor product cubature and still retain the designed exactness within each block.

For each block $\iBlock$, we construct an equal-weight quadrature $Q_\iBlock[\cdot]$ that exactly integrates a set of $r_\iBlock$ basis functions,
$\Phi_\iBlock = \left\{ \fBasis^{(\iBlock)}_k(\vPa_\iBlock) : \Th_\iBlock \mapsto \reals \mid k \in [r_\iBlock] \right\}$,
against the corresponding block distribution $\pQ(\vPa_\iBlock)$.
Given any function $f(\vPa_\iBlock) \in \sSpan(\Phi_\iBlock)$, we have exactly $\nQuad$ evaluation nodes,
$\vPa^{(\iQuad)}_\iBlock \in \Th_\iBlock$ for $\iQuad \in [\nQuad]$, for which
\begin{align}
\label{eq:blockexact}
Q_\iBlock[f] = \frac{1}{\nQuad} \sum_{\iQuad=1}^\nQuad f(\vPa^{(\iQuad)}_\iBlock)
 = \int f(\vPa_\iBlock)\,d\pQ(\vPa_\iBlock).
\end{align}
For example, a good choice would be a minimal sigma point rule \cite{Wan2007}, based on the simplex-polytope vertices as constructed in \Cref{alg:simplex},
but the higher-order rules mentioned earlier \cite{Smolyak1963,mcnamee1967construction,Gerstner1998,Novak1999,Petras2003}
may also be feasible when limited to sufficiently small blocks.

We can then form a stochastic blocked mean-field quadrature by concatenating independent uniformly-random permutations,
represented by permutation matrices
$P_\iBlock$ for each $\iBlock \in [\nBlock]$, applied to the evaluation nodes in each block as
\begin{align}
\label{eq:equalweight}
\begin{bmatrix}
\vPa^{(1)} & \cdots & \vPa^{(\nQuad)}
\end{bmatrix} = 
\begin{bmatrix}
[ \vPa^{(1)}_1 & \vPa^{(2)}_1 & \cdots & \vPa^{(\nQuad)}_1] P_1 \\
[ \vPa^{(1)}_2 & \vPa^{(2)}_2 & \cdots & \vPa^{(\nQuad)}_2] P_2 \\
 & & \vdots & \\
[ \vPa^{(1)}_\nBlock & \vPa^{(2)}_\nBlock & \cdots & \vPa^{(\nQuad)}_\nBlock] P_\nBlock \\
\end{bmatrix}
\quad\text{so}\quad
Q[f] = \frac{1}{\nQuad} \sum_{\iQuad=1}^\nQuad f(\vPa^{(\iQuad)}).
\end{align}

\begin{theorem}[Expectation Exactness]
\label{thm:expect}
Given a blocked mean-field distribution as described in \Cref{eq:directsum,eq:blockmeanfield},
and a stochastic equal-weight quadrature as in \Cref{eq:blockexact,eq:equalweight},
then for any function that is a product of exact functions within each block,
\begin{align*}
f(\vPa) = \prod_{\iBlock=1}^\nBlock f^{(\iBlock)}(\vPa_\iBlock)\quad\text{where}\quad f^{(\iBlock)}\in \sSpan(\Phi_\iBlock)
\quad\text{for each}\quad
\iBlock \in [\nBlock],
\end{align*}
the expectation of the quadrature is exact, i.e.~
\begin{align*}
\expect_{P_1, P_2, \ldots, P_\nBlock} Q[ f ] = \int f(\vPa)\,d\pQ(\vPa).
\end{align*}
\end{theorem}

See \Cref{sub:pf_expect} for a short proof.
The key idea is that by composing equal-weight quadratures from each block of coordinates with concatenation,
the result retains the same exactness for functions restricted to each block.
Since each quadrature is equal-weight and the evaluation nodes are permuted uniformly at random,
the probability of concatenating any specific sequence of evaluations nodes is a constant that coincides with the weight of each node in the tensor product cubature.

\DeclareDocumentCommand \cA{} {\sqrt{2}}
\DeclareDocumentCommand \cB{} {\frac{-1}{\sqrt{2}}}
\DeclareDocumentCommand \pC{} {\sqrt{\frac{3}{2}}}
\DeclareDocumentCommand \nC{} {-\sqrt{\frac{3}{2}}}

\paragraph{Extra Exactness}
The exactness we obtain from this method goes beyond the partition of consecutive blocks.
To understand this, consider this example of concatenated 2-block sigma points (vertices of an equilateral triangle) in 4 pairs:
\begin{align*}
\begin{bmatrix}
\vPa^{(1)T} \\
\vPa^{(2)T} \\
\vPa^{(3)T} \\
\end{bmatrix}
=
\begin{bmatrix}
\cA & 0   & \cB & \nC & \cA & 0   & \cB & \pC \\
\cB & \nC & \cA & 0   & \cB & \pC & \cB & \nC \\
\cB & \pC & \cB & \pC & \cB & \nC & \cA & 0   \\
\end{bmatrix}.
\end{align*}
As intended, these evaluation nodes contain sigma points in the 1-2 block, as well as the 3-4, 5-6, and 7-8 blocks.
However, we also obtained sigma points in the 1-6 and 2-5 blocks, since the permutations happen to have produced compatible pairings.

\begin{figure}[ht]
	\centering
	\includegraphics[width=0.90\textwidth]{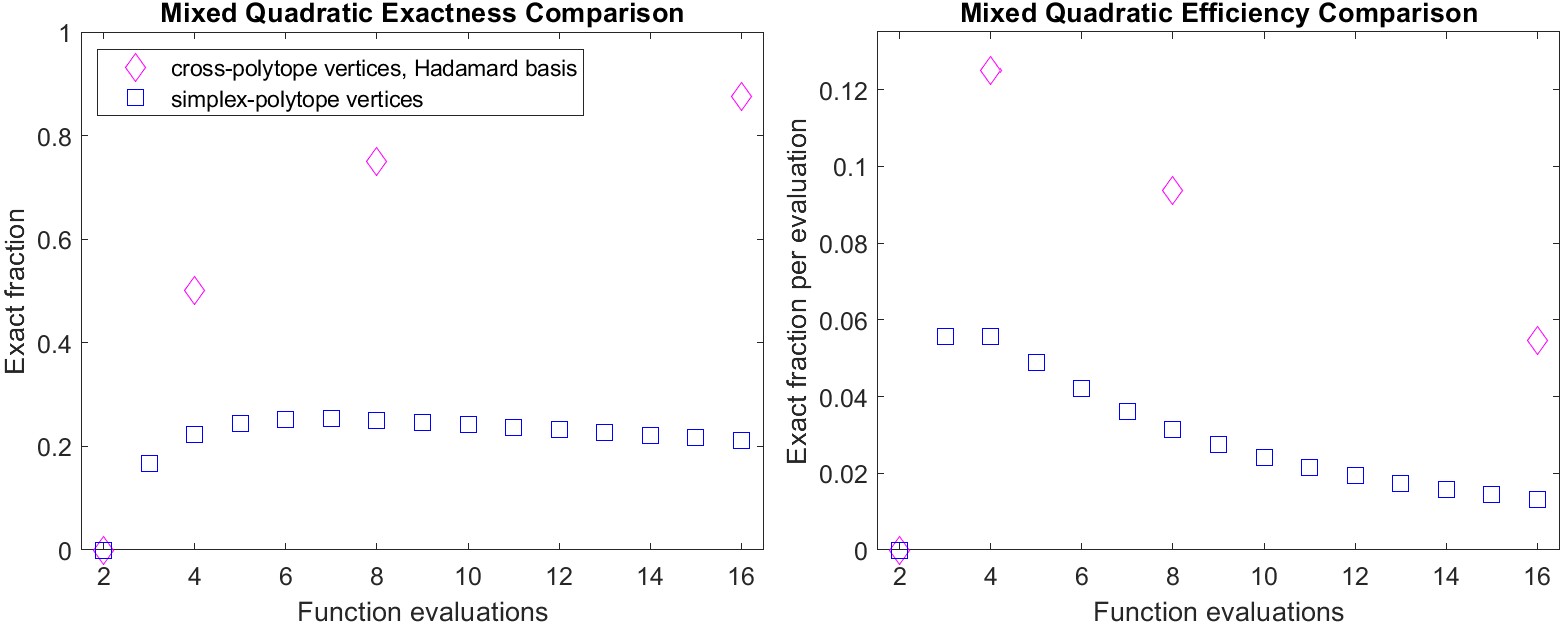}
	\precaption
	\caption{
	\textbf{Left:} Increasing the block dimension for the simplex-based quadratures does not necessarily increase the number of exact mixed quadratic basis functions.
	\textbf{Right:} Using a block size of 2 or 3 for the simplex-based quadratures (which require 3 or 4 function evaluations, respectively) will maximize the number of exact basis functions per evaluation.
	However, the cross-polytope sequence is even more efficient, with the optimum at 4 function evaluations per quadrature.
	}
	\postcaption
	\label{fig:extra_exact}
\end{figure}

\Cref{fig:extra_exact} shows how extra exactness for mixed quadratic basis functions varies with the block size.
Within each block, the quadrature uses sigma points composed of the simplex-polytope vertices.
Since larger block sizes increase the number of permutations, the probability of realizing compatible permutations drops for larger blocks
and the reduction in extra exactness can outweigh the increased exactness within each block.
This comparison also includes quadratures from \Cref{sub:quasirandom} (i.e.~cross-polytope vertices in the Hadamard basis), which are discussed next.
Each numerical experiment counts exact off-diagonal integrals on a $6000 \times 6000$ mean-field covariance matrix, which should all be zero,
and the average over 100 trials is shown.

\subsection{Cross-Polytope Vertices in the Hadamard Basis}
\label{sub:quasirandom}

Unfortunately, we often encounter a substantial difference between the integration error corresponding to the expectation,
i.e.~the limit of averaging samples over increasingly long sequences,
and the error corresponding to a single quadrature or even an average of a few samples.
What follows is the result of trying to develop a quasirandom sequence to suppress error more efficiently.

Consider concatenating a set of two-point equal-weight quadratures in each parameter dimension.
This is the minimum number of evaluation nodes required to exactly integrate all univariate quadratic functions,
thus recovering the primary contributions to both the mean and diagonal covariance of a mean-field distribution.
Note, however, that errors will persist due to off-diagonal covariance integrals that do not correctly vanish, e.g.~$\expect_{\vQ}[(\vPa_1 - \vMu_1)(\vPa_2 -\vMu_2)]=0$.
Rather than averaging several quadratures composed by random permutations, we could just form tensor product cubatures in adjacent pairs.
If the mean is $\expect_{\vQ} \left[ \vPa \right] = \vMu$ and $\expect_{\vQ} \left[ (\vPa-\vMu)(\vPa-\vMu)^T \right] = \diag(\vSigma^2)$, then these four evaluation nodes are
\begin{align*}
\begin{bmatrix}
 \vTh^{(1)} & \vTh^{(2)} & \vTh^{(3)} & \vTh^{(4)} \\
\end{bmatrix} = \vMu + \diag(\vSigma)
\begin{bmatrix}
 1 & -1 & 1 & -1 \\
 1 & -1 & -1 & 1 \\
 1 & -1 & 1 & -1 \\
 1 & -1 & -1 & 1 \\
&&\vdots&\\
\end{bmatrix}.
\end{align*}
Not only does this result in exact pairwise cubatures with only four function evaluations,
the exact cubature blocks actually include all $\frac{d^2}{4}$ pairs of dimensions containing both an even parameter and an odd parameter.
One can easily show that this is the maximum number of exact blocks that can be obtained by switching some of the signs of the second pair of nodes.

Building on this strategy, we can then partition coordinates into consecutive 4-blocks.
Repeating the evaluation nodes, but flipping signs in alternating 2-blocks, results in a tensor product cubature in each 4-block.
Iterating these node sequences to larger blocks to obtain cubatures of still higher dimensions yields the cross-polytope vertices in the Hadamard basis.
\Cref{alg:cross-polytope} generates the signs needed to construct an antithetic pair of vertices from the iterate index, $\iQuad = 0, 1, \ldots, 2^{\lceil\log_2(\nPa)\rceil}-1$.
Each quadrature from this sequence still exactly integrates all linear combinations of univariate quadratics, or $1 + 2 \nPa$ basis functions, against the mean-field distribution.
If the mean-field distribution is symmetric in each coordinate, all linear combinations of univariate cubics are exact as well.

\begin{figure}[ht]
	\centering
	\includegraphics[width=0.90\textwidth]{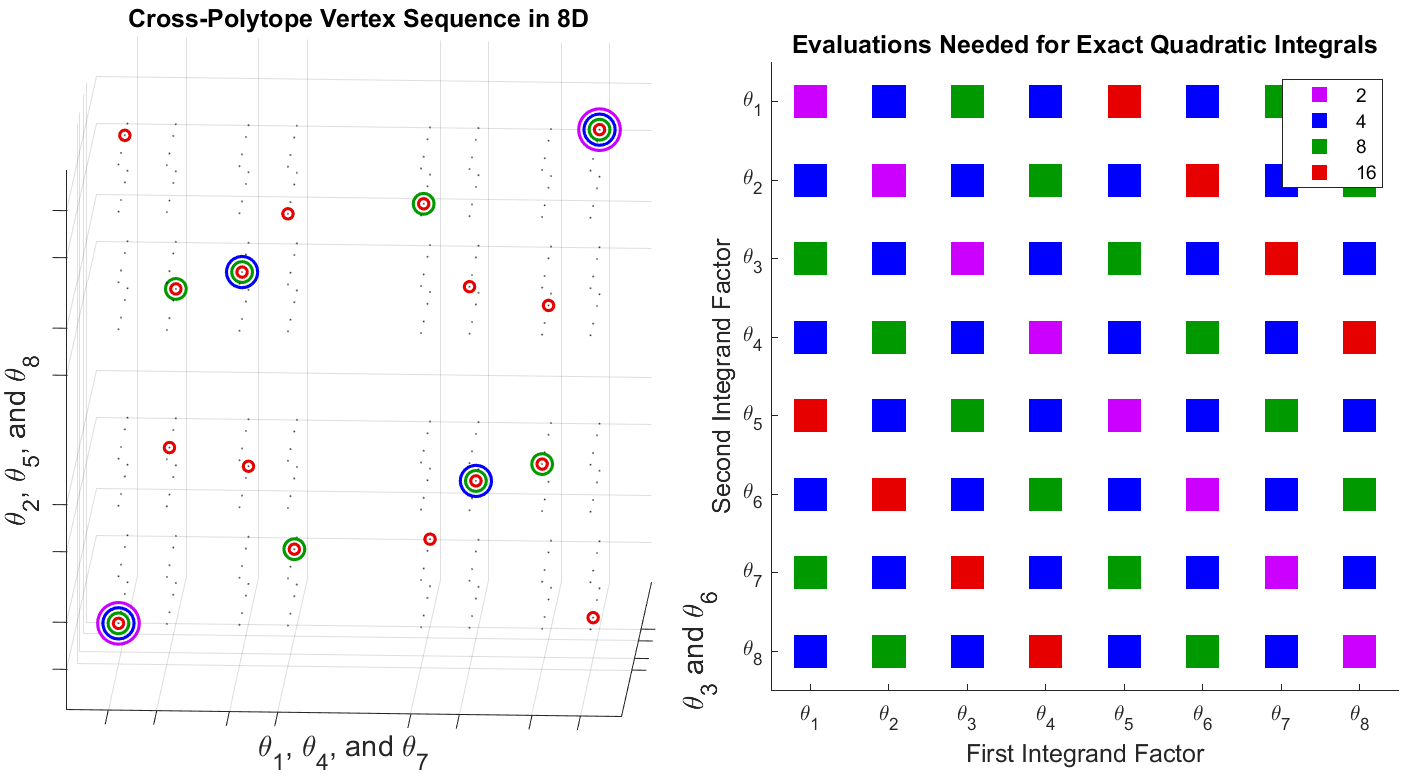}
	\caption{
	\textbf{Left:} Illustration of cross-polytope vertex evaluations in 8D, visualized with the nested coordinate scaling
	$x = \frac{2}{5}(\frac{2}{5} \theta_1 + \theta_4) + \theta_7$,
	$y = \frac{2}{5}(\frac{2}{5} \theta_2 + \theta_5) + \theta_8$, and
	$z = \frac{2}{5} \theta_3 + \theta_6$.
	The fine grid points are hypercube vertices, a superset of the cross-polytope vertices in the Hadamard basis.
	Every antithetic pair, e.g.~in lavender, exactly integrates all univariate quadratics.
	Longer sequences---shown in blue, green, and red---exactly integrate increasing sets of multivariate quadratics. \\
	\textbf{Right:} Visualization of the number of function evaluations from this sequence needed to integrate each product.
	Every four evaluations creates 2D cubatures for half of all coordinate pairs (exact on lavender diagonal and blue checkerboard).
	}
	\label{fig:cross-poly}
\end{figure}

\Cref{lem:relative_parity} and \Cref{thm:exactness_periodicity} show how \Cref{alg:cross-polytope} produces quadrature subsequences that obtain periodic exactness within two-dimensional subspaces.
See \Cref{sub:pf_relative_parity} and \Cref{sub:pf_exactness_periodicity} for proofs.
\Cref{fig:cross-poly} provides a visualization of the cross-polytope sequence in the Hadamard basis (left) and the number of function evaluations needed to correctly integrate specific covariance basis functions (right).

\begin{lemma}[Relative Parity]
\label{lem:relative_parity}
The relative parity, $\vP_{\iPa_1} \xor \vP_{\iPa_2}$, corresponding to any two distinct parameters, $\vPa_{\iPa_1}$ and $\vPa_{\iPa_2}$, in \Cref{alg:cross-polytope}
determines the product of corresponding signs in the result
and it only depends on the iterate index $\iQuad$ and the binary string obtained by bitwise exclusive disjunction of binary representations of parameter indices,
\begin{align*}
&\vP_{\iPa_1} \xor\, \vP_{\iPa_2} = \xor_{\iBit=1}^{\nBit} \left[ \vX_\iBit \and \bit_{\iBit}(\iQuad) \right] \\
&\text{where}\quad
\vX_\iBit = \bit_{\iBit}(\iPa_1) \xor \bit_{\iBit}(\iPa_2) \quad\text{for}\quad \iBit \in [\nBit].
\end{align*}
\end{lemma}

\begin{theorem}[Exactness Periodicity]
\label{thm:exactness_periodicity}
Let $\vPa_{\iPa_1}$ and $\vPa_{\iPa_2}$ be any two distinct parameters.
Let $b \in [\nBit]$ indicate the position of the least-significant bit that is different between both binary representations of their indices, $\iPa_1$ and $\iPa_2$.
Every consecutive contiguous quadrature subsequence of $2^b$ antithetic pairs obtained by \Cref{alg:cross-polytope},
using iterate indices $\iQuad = z 2^b, z 2^b + 1, \ldots, (z+1) 2^b - 1$ for $z \in \mathbb{Z}_{\geq 0}$,
averages to the 4-node tensor-product cubature over the corresponding two-dimensional subspace.
\end{theorem}

\begin{algorithm}[htb]
\caption{Cross-Polytope Vertex Sequence in Hadamard Basis}
\label{alg:cross-polytope}
\fontsize{10}{16}\selectfont
\begin{algorithmic}[1]
\Statex $\nPa$ is the number of parameter dimensions.
\Statex $\iQuad$ is a non-negative integer index for the desired term of the sequence. 
\Ensure $\vS$ is the $\nPa \times 1$ vector of signs needed to construct an antithetic pair, $\vPa^{(2\iQuad+1)} = \vMu + \vS \ast \vSigma$ and $\vPa^{(2\iQuad+2)} = \vMu - \vS \ast \vSigma$,
for an equal-weight ($w = \frac{1}{2}$) quadrature for a distribution with mean $\vMu$ and covariance $\diag(\vSigma)^2$.
\Function{$\vS = \fCrossPolyQuad$}{$\nPa, \iQuad$}
\State Get the number of bits needed to index parameters, $\nBit = \lceil\log_2(\nPa)\rceil$.
\State Compute the parity of each parameter dimension for iterate $\iQuad$,
\begin{align*}
\vP_\iPa = \xor_{\iBit=1}^{\nBit} \left[ \bit_{\iBit}(\iPa) \and \bit_{\iBit}(\iQuad) \right]
\quad\text{for all}\quad
\iPa \in \left\{ 0, 1, \ldots, \nPa-1 \right\}.
\end{align*}
\State Return signs from parity, $\vS = 2 \vP - 1$.
\EndFunction
\end{algorithmic}
\end{algorithm}

These subsequences also generate higher-order cubatures in up to $\nBit+1$ dimensions for some specific sets of parameters.
This occurs when the binary representations of parameter indices only differ by a single bit, each, from a base index.
For example, if parameters $\vPa_{\iPa_1}, \vPa_{\iPa_2}, \vPa_{\iPa_3},$ and $\vPa_{\iPa_4}$ are such that only
$\bit_1(\iPa_1) \neq \bit_1(\iPa_4)$,
$\bit_2(\iPa_2) \neq \bit_2(\iPa_4)$,
$\bit_3(\iPa_3) \neq \bit_3(\iPa_4)$,
and all other bits are the same as $\iPa_4$,
then every 8 quadratures (16 evaluations) would result in a tensor-product cubature for functions that only depend on these parameters.
This easily follows by applying the same reasoning as used in \Cref{sub:pf_exactness_periodicity}.
Unfortunately, this observation bears little importance since errors in many two-dimensional subspaces will still persist, and dominate, until $2^\nBit$ quadratures have been averaged.

\section{Variational Fixed-Point Optimization}
\label{sec:variational}

Efficient numerical integration enables a simple optimization procedure for mean-field variational inference as a fixed-point iteration.
\Cref{sub:basis_analysis} analyzes the relationship between an optimal variational approximation and the corresponding integrals needed to project the posterior distribution onto a variational basis.
\Cref{sub:quadratic} shows how to modify the preceding quadrature sequences to efficiently implement these posterior projections onto a quadratic basis for Gaussian mean-field distributions.
\Cref{sub:dirac_gauss} provides an extension to mean-field distributions that may support parameter sparsification.
\Cref{sub:qnvb} discusses implementation details for quasi-Newton variational Bayes, an implementation of projective integral updates for Gaussian mean fields.

\subsection{Projective Integral Updates}
\label{sub:basis_analysis}

Let us consider a family of variational distributions that may be written as an exponential of a linear combination of basis functions, $\fBasis_\iBasis( \vPa )$ for $\iBasis=0, 1, \ldots, \nBasis$.
That is,
\begin{align}
\label{eq:varexp}
\pQ(\vPa \mid \vPhi) = \exp\left[ \sum_{\iBasis=0}^\nBasis \vPhi_\iBasis \fBasis_\iBasis( \vPa ) \right]
\quad\text{and}\quad
\vPhi^* = \argmin_{\vPhi}\, \KL{\pQ(\vPa \mid \vPhi)}{\pP(\vPa \mid \sData)}
\end{align}
is the vector of coefficients corresponding to the optimizer.
We can define the basis function that controls normalization as $\fBasis_0(\vPa) \equiv 1$.
Since each variational distribution induces an inner product on functions $f(\vPa)$ and $g(\vPa)$,
\begin{align}
\label{eq:innerproduct}
\left\langle f, g \right\rangle_{\vPhi} = \int f(\vPa) g(\vPa)\,d\pQ(\vPa \mid \vPhi),
\end{align}
we can also construct the basis to be orthogonal, i.e.~$\langle \fBasis_i, \fBasis_j \rangle_{\vPhi^*} = 0$ for all $i \neq j$.
This framework allows us to apply the calculus of variations to illuminate the relationship between optimal coefficients and the posterior distribution $\pP(\vPa \mid \sData)$.
In particular, we can optimize the variational density with projective integral updates: 
\begin{align}
\label{eq:pid}
\vPhi^{(t+1)}_\iBasis = \frac{\left\langle \fBasis_\iBasis, \log\!\left( \pP(\sData \mid \vPa) \pP(\vPa) \right) \right\rangle_{\vPhi^{(t)}}}{\left\langle \fBasis_\iBasis, \fBasis_\iBasis \right\rangle_{\vPhi^{(t)}}}
\quad\text{for}\quad \iBasis = 1, 2, \ldots, \nBasis,
\end{align}
where $\vPhi_0^{(t+1)}$ is determined by normalization.

Analysis proceeds by considering arbitrary infinitesimal perturbations in the vicinity of the optimizer by using a differential element $\eps$ and a vector of perturbations $\vEta$
to write Gateaux derivatives, $\pQ(\vPa \mid \vPhi = \vPhi^* + \eps \vEta)$.
Feasible perturbations are constrained to only those that retain normalization in \Cref{lem:norm_preserving}.
\Cref{thm:projective_integral} follows by applying the variational principle to perturbations about the optimizer.
See \Cref{sub:pf_norm_preserving,sub:projective_integral} for proofs.

\begin{lemma}[Normalization Preserving Perturbations]
\label{lem:norm_preserving}
Given a variational family satisfying \Cref{eq:varexp} with an orthogonal basis under the inner product \cref{eq:innerproduct},
infinitesimal normalization-preserving perturbations $\vEta$ about any feasible $\vPhi$ must satisfy $\vEta_0 = 0$.
\end{lemma}

\begin{theorem}[Projective Integral Fixed-Point]
\label{thm:projective_integral}
Under the condition of \Cref{eq:varexp} and using the inner product defined by \Cref{eq:innerproduct},
the optimal coefficients $\vPhi^*$ are a fixed point of \Cref{eq:pid},
projecting the log-posterior onto the span of the variational basis and normalizing the result.
\end{theorem}

\subsection{Gradient and Hessian Extraction}
\label{sub:quadratic}

Perhaps the simplest mean-field distribution that is amenable to this approach uses a quadratic basis in each coordinate, yielding a mean-field Gaussian.
Since we typically think about minimizing loss during optimization of learning algorithms, we will frame analysis in terms of the negative log-posterior.
At time $t$, let $\vMu^{(t)}$ be the expansion point for a quadratic basis,
$\zeroLoss$ is the constant offset, $\vG$ is the average gradient, $\vH$ is the average Hessian diagonal, and the residual $r(\theta)$ contains all other terms so that the loss is
\begin{align*}
\fLoss(\vPa \mid \sData) &= -\log\left( \pP(\sData \mid \vPa)\pP(\vPa) \right) \\
&= \zeroLoss + \left(\vPa - \vMu^{(t)}\right)^T \left[\vG + \frac{1}{2} \vH \ast \left(\vPa - \vMu^{(t)}\right) \right] - \fResidual(\vPa).
\end{align*}
To maintain stability and normalizability, the Hessian diagonal must be constrained as $\vHh_\iPa = \max(\vH_\iPa,\, h_\text{min})$ for some $h_\text{min} > 0$,
allowing us to solve for the mean update, $\vMu^{(t+1)} = \vMu^{(t)} - \vG \ast \vHh^{-1}$, and standard deviation, $\vSigma^{(t+1)}_\iPa = \vHh_\iPa^{-\nicefrac{1}{2}}$, to obtain
\begin{align*}
\pQ(\vPa \mid \vPhi^{(t+1)}) = \pN( \vTh \mid \vMu^{(t+1)}, \diag(\vSigma^{(t+1)2})),
\end{align*}
where each $\vPhi^{(t+1)}$ contains both $\vMu^{(t+1)}$ and $\vSigma^{(t+1)}$.
Thus, we recover quasi-Newton steps from averaged, rather than instantaneous, gradients and Hessian diagonals.

In principle, we could approximate the gradient and Hessian expectations by evaluating orthogonal projections directly.
Dropping time superscripts, the relevant projection coefficients are
\begin{align}
\vG_i = \vSigma_\iPa^{-2} \left\langle \vPa_\iPa - \vMu_\iPa,\,\fLoss(\vPa \mid \sData) \right\rangle_{\vPhi} \quad\text{and}\quad \label{eq:gint}\\
\vH_i = \vSigma_\iPa^{-4} \left\langle \left(\vPa_\iPa - \vMu_\iPa\right)^2 - \vSigma_\iPa^{2},\, \fLoss(\vPa \mid \sData) \right\rangle_{\vPhi},\label{eq:hint}
\end{align}
but a much more efficient approach leverages loss gradients.

A quadrature may be generally understood as a linear combination of linear functionals (e.g.~point-wise evaluations) that approximates another linear functional (e.g.~integration against the mean-field distribution).
If the approximation is exact for some basis, then the quadrature must also be exact for the entire span by applying linearity.
While the quadratures we have considered so far employ point-wise evaluations of a function,
point-wise evaluations of the gradient also comprise linear functionals, $\nPa$ of them,
and allow us to evaluate many basis coefficients simultaneously.
Since we can write each basis function in \Cref{eq:gint,eq:hint} as derivatives,
\begin{align*}
& \frac{\partial}{\partial \vPa_\iPa} \pN( \vPa_\iPa \mid \vMu_i,\, \vSigma_i ) = -\vSigma_\iPa^{-2} (\vPa_\iPa - \vMu_\iPa) \pN( \vPa_\iPa \mid \vMu_i,\, \vSigma_i^2 )\quad\text{and} \\
& \frac{\partial}{\partial \vPa_\iPa} (\vPa_\iPa - \vMu_\iPa) \pN( \vPa_\iPa \mid \vMu_i,\, \vSigma_i ) = -\vSigma_\iPa^{-2} \left( (\vPa_\iPa - \vMu_\iPa)^2 - \vSigma_\iPa^2 \right) \pN( \vPa_\iPa \mid \vMu_i ,\, \vSigma_i )\,,
\end{align*}
respectively, integrating by parts casts these derivatives onto the loss,
enabling simultaneous projections for all parameters from the gradient,
\begin{align}
& \vG = \int \nabla_{\vPa} \fLoss(\vPa \mid \sData) \,d\pN( \vPa \mid \vMu,\, \diag(\vSigma^2) ) \quad\text{and} \label{eq:ibpg}\\
& \vH = \vSigma^{-2} \ast \int (\vPa - \vMu) \ast \nabla_{\vPa} \fLoss(\vPa \mid \sData) \,d\pN( \vPa \mid \vMu,\, \diag(\vSigma^2) )
\,.\label{eq:ibph}
\end{align}

\paragraph{Relationship to Hutchinson's Method}
Given the loss gradient, one can generate a random variable $\vZ$
and form the inner product, $\vZ^T \nabla_{\vPa} \fLoss(\vPa \mid \sData)$.
Taking a second gradient gives $\nabla_{\vPa} \left( \vZ^T \nabla_{\vPa} \fLoss(\vPa \mid \sData) \right) = \mH \vZ$,
where $\mH$ is the full Hessian matrix at $\vPa$.
If $\vZ$ has mean zero and identity covariance, it follows that $\expect[ \vZ \ast \mH \vZ ] = \diag(\mH)$ \cite{bekas2007estimator}.
When we apply a quadrature to \Cref{eq:ibph}, we obtain a finite difference analogue that uses multiple gradient evaluations in exchange for a second stage of backpropagation,
and the result approximates the expected Hessian, rather than the instantaneous Hessian.
\Cref{alg:quad-loss} shows how to implement this approach with the quadrature sequence from \Cref{sub:quasirandom}
to construct a local quadratic approximation of the loss from gradients.
 
\begin{algorithm}[htb]
\caption{Quadratic Loss Approximation}
\label{alg:quad-loss}
\fontsize{10}{16}\selectfont
\begin{algorithmic}[1]
\Require $\vMu$ is a $\nPa \times 1$ vector of means, where $\nPa$ is the parameter dimension.
\Statex $\vSigma$ is a $\nPa \times 1$ vector of standard deviations of the mean-field distribution.
\Statex $\iQuad_1$ is an integer indicating the next position within the quadrature sequence.
\Statex $\nQuad$ indicates the number of antithetic pairs to use from the quadrature sequence.
\Statex $\fLoss(\cdot)$ is a loss function, returning both the value and gradient.   
\Ensure $\zeroLoss$, $\vG$, and $\vH$ so that $\fLoss(\vPa) \approx \zeroLoss + (\vPa - \vMu)^T \vG + \frac{1}{2} (\vPa - \vMu)^T \diag(\vH) (\vPa - \vMu)$.
\Function{$(\zeroLoss,\, \vG,\, \vH) = \fQuadraticApprox$}{$\vMu, \vSigma, \iQuad_1, \nQuad, \fLoss$}
\State $\zeroLoss = 0$; \quad $\vG = 0^{\nPa \times 1}$; \quad $\vH = 0^{\nPa \times 1}$
\Comment{Initialize accumulators.}
\For{$\iQuad = \iQuad_1, \iQuad_1+1, \ldots, \iQuad_1 + \nQuad - 1$}
	\State $\vS = \fCrossPolyQuad(\nPa, k)$
	\Comment{Get quadrature signs.}
	\State $[\zeroLoss_{+},\, \vG_{+}] = \fLoss( \vMu + \vSigma\ast\vS )$
	\State $[\zeroLoss_{-},\, \vG_{-}] = \fLoss( \vMu - \vSigma\ast\vS )$
	\State $\zeroLoss \gets \zeroLoss + \zeroLoss_{+} + \zeroLoss_{-}$; \quad $\vG \gets \vG + \vG_{+} + \vG_{-}$
	\Comment{Ordinary integration.}
	\State $\vH \gets \vH + (\vG_{+} - \vG_{-})\ast\vS$
	\Comment{Integrate product against first-order basis.}
\EndFor
\State $\vG \gets \frac{1}{2\nQuad} \vG$
\State $\vH \gets \vH \ast (2\nQuad \vSigma)^{-1}$
\State $\zeroLoss \gets \frac{\zeroLoss}{2\nQuad} - \frac{\vH^T \vSigma^2}{2}$
\EndFunction
\end{algorithmic}
\end{algorithm}

\subsection{Dirac-Gauss Mixtures}
\label{sub:dirac_gauss}

Now we will consider a mean-field framework that is capable of capturing a finite probability that any individual coordinate $\vTh_i$ is zero.
By introducing a Bernoulli-distributed random variable $\vZ_i$ we can write each mean-field factor as Dirac-Gauss mixture by marginalizing over $\vZ_i$ as
\begin{align}
\label{eq:dirac_gauss}
\pQ(\vTh_i) &= \sum_{\vZ_i \in \{0, 1\}} \pQ(\vTh_i, \vZ_i) = \sum_{\vZ_i \in \{0, 1\}} \pQ(\vZ_i) \pQ(\vTh_i \mid \vZ_i) \\
&= \pQ(\vZ_i) \pD_\eps(\vTh_i) + \pQ(\lnot \vZ_i)\pN_\eps(\vTh_i \mid \vNu_i, \vTau_i^2).\nonumber
\end{align}
This expression uses shorthand, $\pQ(\vZ_i) \equiv \pQ(\vZ_i=1)$ and $\pQ(\lnot\vZ_i) \equiv \pQ(\vZ_i=0)$,
and conditional distributions can be constructed to be non-overlapping by carving out a small interval, $( -\frac{\eps}{2}, \frac{\eps}{2} )$ for some $\eps > 0$,
\begin{align*}
& \pD_\eps(\vTh_i) =
\begin{cases}
\eps^{-1} & |\vTh_i| < \frac{\eps}{2} \\
0       & \tElse \\
\end{cases}
\quad\text{and}\quad
\pN_\eps(\vTh_i \mid \vNu_i, \vTau_i^2) =
\begin{cases}
0 & |\vTh_i| < \frac{\eps}{2} \\
\pN(\vTh_i \mid \vNu_i, \vTau_i^2)       & \tElse \\
\end{cases}.
\end{align*}
In the limit $\eps \rightarrow 0$, exact normalization of $\pN_\eps(\vTh_i \mid \vNu_i, \vTau_i^2)$ is unnecessary,
because doing so only multiplies the scaling factor by $1 + \mathcal{O}(\eps)$, a vanishing change.
We can also construct a spike-and-slab prior with the same structure,
\begin{align*}
\pP(\vTh_i) = \pP(\vZ_i) \pD_\eps(\vTh_i) + \pP(\lnot \vZ_i)\pN_\eps(\vTh_i \mid 0, \hPr^{\!\!-1}),
\end{align*}
where $\hPr$ is the prior precision associated with a nonzero.

Note the important distinction between the mean and standard deviations required by the quadrature formulas in \Cref{alg:quad-loss}
versus the mean and variance of the normal distribution in \Cref{eq:dirac_gauss}, which is conditioned on a nonzero.
The correct moments of this distribution are:
\begin{align}
\label{eq:dirac_gauss_moments}
\vMu_i = \pQ(\lnot \vZ_i) \vNu_i
\quad\text{and}\quad
\vSigma_i^2 = \pQ(\vZ_i)\pQ(\lnot\vZ_i) \vNu_i^2 + \pQ(\lnot\vZ_i) \vTau_i^2.
\end{align} 

To apply the analysis from \Cref{sub:basis_analysis}, we need to write the variational family as an exponential,
so it is useful to formulate the probabilities of zeros as logits:
\begin{align*}
\logitPr &= \log\!\left( \frac{\pP(\vZ_i)}{\pP(\lnot\vZ_i)} \right),
& \pP(\vZ_i) &= \frac{\exp( \logitPr )}{1 + \exp( \logitPr )}, \\
\vZeta_i &= \log\!\left( \frac{\pQ(\vZ_i)}{\pQ(\lnot\vZ_i)} \right),
& \text{and}\quad
\pQ(\vZ_i) &= \frac{\exp( \vZeta_i )}{1 + \exp( \vZeta_i )}.
\end{align*}
Since the conditional distributions are disjoint, we can easily compute
\begin{align}
\label{eq:log_p}
\log \pP(\vTh_i) &= -\log\left( 1 + \exp( \logitPr ) \right) + 
\begin{cases}
\logitPr - \log(\eps) & |\vTh_i| < \frac{\eps}{2} \\
\frac{1}{2}\log( \frac{\hPr}{2\pi} ) - \frac{1}{2} \hPr \vTh_i^2 & \tElse \\
\end{cases} \\
&= \tConst - \frac{\hPr \vTh_i^2}{2}  +  
\begin{cases}
\logitPr + \frac{1}{2}\log( \frac{2\pi}{\hPr} ) - \log(\eps) & |\vTh_i| < \frac{\eps}{2} \\
0 & \tElse \\
\end{cases} \nonumber.
\end{align}
Likewise, each variational factor can be written
\begin{align}
\label{eq:log_q}
\log \pQ(\vTh_i) = \tConst - \frac{(\vTh_i - \vNu_i)^2}{2\vTau_i^2} +  
\begin{cases}
\vZeta_i + \frac{1}{2}\log\left( 2\pi\vTau_i^2 \right) + \frac{\vNu_i^2}{2\vTau_i^2} - \log(\eps) & |\vTh_i| < \frac{\eps}{2} \\
0 & \tElse \\
\end{cases}.
\end{align}
Since \Cref{alg:quad-loss} allows us to project the loss onto a quadratic univariate basis,
we only need to add the prior discontinuity term in \Cref{eq:log_p}
to continuous quadratic projections of the remaining log posterior components.
For example, adding the quadratic terms from the negative log-prior and negative log-likelihood, with gradient $\vG$ and Hessian diagonal $\vH$ at the expansion point $\vMu^{(t)}$,
gives
\begin{align*}
&\frac{\hPr}{2} \vTh_i^2 + \vG_i(\vTh_i - \vMu^{(t)}) + \frac{\vH_i}{2}  (\vTh_i - \vMu^{(t)})^2
= \tConst + \frac{(\vTh_i - \vNu_i)^2}{2\vTau_i^2} \\
&\text{where}\quad
\vNu_i = \frac{\vH_i \vMu^{(t)} - \vG_i}{\hPr + \vH_i}\quad\text{and}\quad
\vTau_i^2 = (\hPr + \vH_i)^{-1}.
\end{align*}
After absorbing constants into the residual, the remaining prior terms give
\begin{align}
\label{eq:log_post}
\log \pP(\vTh \mid \sData) &= \fResidual(\vPa) + \sum_\iPa \left( \frac{-(\vTh_i - \vNu_i)^2}{2\vTau_i^2}
+\begin{cases}
\logitPr + \frac{1}{2}\log\!\left( \frac{2\pi}{\hPr} \right) - \log(\eps) & |\vTh_i| < \frac{\eps}{2} \\
0 & \tElse \\
\end{cases}\right).
\end{align}
Matching \Cref{eq:log_post} to \Cref{eq:log_q} gives the logit of the zero probability,
\begin{align}
\label{eq:zeta}
\vZeta_i = \logitPr + \frac{1}{2}\left[\log\!\left(\frac{\hPr + \vH_i}{\hPr}\right) - (\hPr + \vH_i) \vNu_i^2 \right].
\end{align}

\subsection{Quasi-Newton Variational Bayes}
\label{sub:qnvb}

\newVector<\bG>{\bar{g}}
\newVector<\bH>{\bar{h}}
\newVector<\bS>{\bar{s}}
\newScalar<\sigmaMin>{\sigma_\text{min}}
\newScalar<\sigmaMax>{\sigma_\text{max}}
\newScalar<\sMin>{s_\text{min}}
\newScalar<\sMax>{s_\text{max}}

Here we examine practical measures used in QNVB to implement robust projective integral updates for Gaussian mean fields.
To design an algorithm that competes with Adam,
we must incorporate safety measures to keep updates to the mean within a controllable trust region.
This is particularly important for parameter dimensions that are not sufficiently constrained by the Hessian, which affects the following updates:
\begin{enumerate}
\item the quasi-Newton step, $\vMu \gets \vMu - \vDelta$ where $\vDelta = \vH^{-1} \ast \vG$,
\item the update to standard deviations, $\vSigma = \vH^{-\nicefrac{1}{2}}$, and
\item the update to the gradient\footnote{%
If we can take full quasi-Newton steps, then the updated gradient is obviously just zero,
but when these steps are restricted (owing to small Hessian diagonals),
this update ensures that the gradient remains consistent with our local quadratic loss approximation.}, $\vG \gets \vG - \vH \ast \vDelta$.
\end{enumerate}
To understand the measures that safeguard these computations, we briefly review some standard practices in machine learning training.

\paragraph{Exponentially-Damped Averages}
Training algorithms based on momentum typically track a running exponentially-damped average of the gradients.
Adam also includes a second average that tracks the raw second moments as
\begin{align}
& \bG^{(t)} = (1 - \beta_1^t)^{-1} (1 - \beta_1) \sum_{j=1}^{t} \beta_1^{t-j} \vG^{(j)} \quad\text{and} \\
& \bS^{(t)} = (1 - \beta_2^t)^{-1} (1 - \beta_2) \sum_{j=1}^{t} \beta_2^{t-j} \vG^{(j)2}.
\end{align}
The default damping coefficients are $\beta_1=0.9$ and $\beta_2=0.999$, corresponding to effective target sample sizes of $\tau_1=(1-\beta_1)^{-1}=10$ and $\tau_2=(1-\beta_2)^{-1}=1000$, respectively.
Because the gradients change quickly, a small sample size produces a more current average.
In contrast, the second moment is more persistent and benefits from more samples to increase precision and stability.
Similarly, the Hessian diagonal is less susceptible to higher-order deviations from a local quadratic loss approximation,
so we can use the same second-order damping coefficient to track the average Hessian diagonal.
Following the method by Yao et al.~\cite{yao2021adahessian}, negative Hessian diagonals can be avoided by tracking the average diagonal-squared,
\begin{align}
\bH^{(t)2} = (1 - \beta_2^t)^{-1} \left( (1 - \beta_2) \sum_{j=1}^{t} \beta_2^{t-j} \vH^{(j)2} \right).
\end{align}

Note that we can also avoid the bias-corrections, $(1 - \beta^t)^{-1}$, if we track the number of effective samples in each running average.
For example, if $\bG^{(t)}$ is an average with $n_1^{(t)}$ effective samples,
then the next gradient evaluation will allow us to update the average to include $n_1^{(t+1)} = \min( n_1^{(t)} + 1, \tau_1 )$ effective samples as
\begin{align}
\hat{\beta}_1 = \frac{n_1^{(t+1)} - 1}{n_1^{(t+1)}}
\quad\text{and}\quad
\bG^{(t+1)} = \hat{\beta}_1 \bG^{(t)} + (1 - \hat{\beta}_1) \vG^{(t+1)}.
\end{align}
When we reach the target, $n_1 = \tau_1$, we recover $\hat{\beta}_1=\beta_1$.
Starting with $n_1^{(0)}=0$, the values in $\bG^{(t)}$ are always correctly scaled, the weight of the newest gradient contribution is $\nicefrac{1}{n_1^{(t)}}$, and the previous terms are correctly reduced to the complement.
This minor modification simplifies the updates in \Cref{alg:step}.

\paragraph{Mean Updates}
The adaptive steps in Adam can be derived by constructing parameter updates that are proportional to the gradient, $\vDelta = \vX \ast \vG$.
By setting the expected perturbation-squared to the learning rate-squared, we have 
\begin{align}
\expect_{\vG}\left[ \vDelta^2 \right] = \lambda^2,
\quad\text{which gives}\quad
\vDelta = \lambda (\bS^{\nicefrac{1}{2}} + \eps)^{-1} \ast \bG.
\end{align}
The stabilizing coefficient $\eps$ places an upper bound on the magnitude of the result when elements of $\bS$ are small.  
Thus, the learning rate serves as an expected perturbation size.
We can use the same technique to enforce a safe upper limit on mean perturbations by simply taking the more restrictive step,
\begin{align}
\vDelta = \min\left( \bH^{(t)\,-1},\, \lambda (\bS^{(t)\,\nicefrac{1}{2}} + \eps)^{-1} \right) \ast \bG^{(t)}
\quad\text{and}\quad
\vMu^{(t)} = \vMu^{(t-1)} - \vDelta.
\end{align}

Since the learning rate only serves as an upper bound to control large parameter perturbations,
we can afford to use a larger learning rate than would be possible with unaltered Adam steps.
This is because an unstable sequence of parameter updates occurs when each step produces a new gradient of larger magnitude than that of the previous step.
Following this reasoning, it is simple to show that a fixed learning rate must be less than twice the inverse of the largest Hessian eigenvalue.
Because we take quasi-Newton steps in the most constrained parameter dimensions (i.e.~those with large Hessian contributions),
the learning rate only affects the least-constrained parameters, leading to a larger stable upper bound.

\begin{algorithm}[b!]
\caption{Variational Step}
\label{alg:step}
\fontsize{10}{16}\selectfont
\begin{algorithmic}[1]
\Require $\fLoss(\cdot)$ is the loss function for the next training case.
\Statex $\vMu$ is a $\nPa \times 1$ vector of means, where $\nPa$ is the parameter dimension.
\Statex $\vSigma$ is a $\nPa \times 1$ vector of standard deviations of the mean-field distribution.
\Statex $\tau_1$ and $\tau_2$ are the target samples computed from $\beta_1$ and $\beta_2$ coefficients.
\Statex $\eps$ is the stabilizing denominator coefficient.
\Statex $\lambda$ is the maximum learning rate.
\Statex $n_1$ and $n_2$ track current samples for running averages in $\bG$, $\bS$, and $\bH^2$.
\Statex $w$ is the annealed weight of cases used in the variational update.    
\Ensure $\vMu$ and $\vSigma$ are updated with the new loss contribution.
\Statex $n_1$, $n_2$, $\bG$, $\bS$, and $\bH$ are also updated.
\Function{$\fStep$}{$\fLoss, \vMu, \vSigma, \tau_1, \tau_2, \eps, \lambda, n_1, n_2, w, \bG, \bS, \bH$}
\State $(\zeroLoss,\, \vG,\, \vH) = \fQuadraticApprox(\vMu, \vSigma, \fLoss(\cdot))$
\State $n_1 \gets \min(n_1 + 1, \tau_1)$ \quad and \quad $\hat{\beta}_1 = \frac{n_1 - 1}{n_1}$.
\State $n_2 \gets \min(n_2 + 1, \tau_2)$ \quad and \quad $\hat{\beta}_2 = \frac{n_2 - 1}{n_2}$.
\State $\bG \gets \hat{\beta}_1 \bG + (1 - \hat{\beta}_1) \vG$
\State $\bS \gets \hat{\beta}_2 \bS + (1 - \hat{\beta}_2) \vG^2$
\State $\bH^2 \gets \hat{\beta}_2 \bH^2 + (1 - \hat{\beta}_2) \vH^2$
\State $\vDelta = \min\left( \bH^{-1},\, \lambda (\bS^{\nicefrac{1}{2}} + \eps)^{-1} \right) \ast \bG$
\State $\vMu \gets \vMu - \vDelta$
\State $\vSigma \gets \max\left(\sigmaMin,\, \max\left(\sMin \vSigma,\, \min\left(\sMax \vSigma,\, \min\left(\sigmaMax,\, (w \bH)^{\nicefrac{-1}{2}} \right)\right)\right)\right)$
\State $\bG \gets \bG - \bH \ast \vDelta$
\EndFunction
\end{algorithmic}
\end{algorithm}

\paragraph{Standard Deviation Updates}
Next, we address safe updates to the standard deviations, which affect the quadrature scales used to compute expectations.
Annealing (tempering) is common practice to support exploration of the posterior in Bayesian inference algorithms.
For example, posterior annealing is formulated as
\begin{align}
\pP(\vPa \mid \sData,\,\alpha) \propto \pP(\sData \mid \vPa)^\alpha \pP(\vPa)^\alpha
\quad\text{where}\quad
0 < \alpha \leq 1.
\end{align}
When applied to variational inference \cite{Mandt2016}, this is equivalent to simply rescaling the total loss, $\fLoss(\vPa \mid \sData,\, \alpha) = \alpha \fLoss(\vPa \mid \sData)$.
Since we track the average gradient and Hessian, annealing can be formulated as a weight $w = \alpha \nData$ that scales the average Hessian by the effective number of cases in the dataset, giving $\vSigma = (w \bH)^{-\nicefrac{1}{2}}$.
QNVB also places absolute bounds, $\sigmaMin$ and $\sigmaMax$, as well as relative bounds, $\sMin$ and $\sMax$, on updates to the standard deviation to avoid rapid changes in scale:
\begin{align}
\vSigma^{(t)} = \max\!\left(\!\sigmaMin,\, \max\!\left(\!\sMin \vSigma^{(t-1)},\, \min\!\left(\!\sMax \vSigma^{(t-1)},\, \min\!\left(\!\sigmaMax,\, (w \bH)^{\nicefrac{-1}{2}} \right)\!\right)\!\right)\!\right)\!.
\end{align}
Reasonable choices of absolute bounds depend on the architecture, but the default relative bounds, $\sMin=0.99$ and $\sMax=1.01$, are fairly robust.

\paragraph{Gradient Updates}
Finally, we can update the running average gradient with the running average Hessian, $\bG^{(t)} \gets \bG^{(t)} - \bH^{(t)} \ast \vDelta^{(t)}$,
which reduces sensitivity to the learning rate because the minimum would not move if we did not also have exponentially-damped averaging.
As long as the full quasi-Newton step can be reached within a few Adam steps, the learning rate makes little difference and only serves to contain the low-curvature dimensions.
All these mechanisms are summarized in \Cref{alg:step}.

\section{Experiments and Discussion}
\label{sec:experiments}

In this section, we examine different training methods for:
1.~image classification with ResNet18 \cite{he2016deep} and CIFAR-10 \cite{Krizhevsky2009},
2.~natural language processing using the Tensorized Transformer \cite{ma2019tensorized} and Penn Treebank (PTB) dataset \cite{marcus1993building}, and
3.~recommendation systems with the deep learning recommendation model (DLRM) \cite{naumov2019deep} and the Criteo Ad Kaggle dataset.

The process used to tune QNVB in these experiments generally consists of a sequence of line searches on $\sigmaMax$, $\lambda$, $\sigmaMin$, and $w$.
Although there are more sophisticated hyperparameter searches, this approach achieves good results with a minimal set of experiments and computational costs. 
Starting with successful Adam hyperparameters ($\lambda$, $\beta_1$, $\beta_2$, $\eps$), a reasonable initialization for $\sigmaMax$ is the same as the learning rate $\lambda$
and $\sigmaMin$ can be set to $10^{-2}$ or $10^{-3}$ times smaller.

It is useful to begin the line search sequence with $\sigmaMax$ because this determines a successful averaging length scale for the least-constrained parameters.
Using several random seeds for each value, a grid search on equally-spaced values in the exponent quickly determines a reasonable magnitude by selecting the value that consistently gives good outcomes.
This process is repeated for the other hyperparameters.

Early experiments showed that $\sMin = 0.99$ and $\sMax = 1.01$ are robust choices and these hyperparameters were not tuned further.
Likewise, typical values for the other Adam parameters ($\beta_1 = 0.9$, $\beta_2 = 0.999$, and $\eps = 10^{-8}$) were not tuned from the published defaults for each architecture.

\subsection{Quadrature Comparisons}
\label{sub:quad}

\paragraph{ResNet18 / CIFAR-10}
QNVB tests for this architecture use 4 gradient evaluations per step from the cross-polytope vertex sequence.
The resulting hyperparameters are $\sigmaMin = 1\times10^{-3}$, $\sigmaMax = 5\times10^{-2}$, $\lambda = 5\times 10^{-3}$, and $w = 4\times10^{4}$.
This likelihood weight is also the size of the full training dataset.
Each training run includes 80 epochs (passes over the full training dataset) and a learning rate schedule that reduces the learning rate by a factor of $1.05$ with each epoch.
This results in a total reduction by a factor of $50$ during training.

\begin{figure}[ht]
	\centering
	\includegraphics[width=0.95\textwidth]{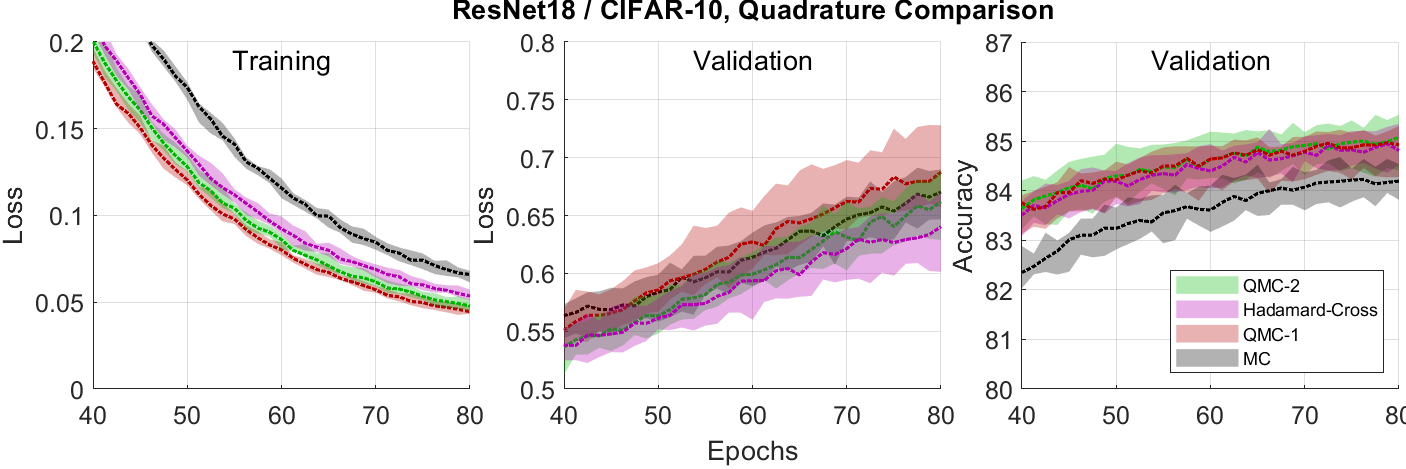}
	\precaption
	\caption{
	\textbf{Left:} Training loss, zoomed to highlight differences during the second half of training.
	These trajectories show that all quadratures that exactly integrate first-order basis functions descend more quickly than Monte Carlo,
	with a significant difference in outcomes appearing early on.
	Note, however, that these outcomes are not reflected by the validation set.\\
	\textbf{Middle:} Validation loss, zoomed to second half of training.
	The validation loss shows that the Hadamard cross-polytope performs somewhat better than the other methods, but there is strong overlap
	with QMC-2. Surprisingly, QMC-1 yields worse validation loss than Monte Carlo.\\
	\textbf{Right:} Validation accuracy during second half of training.
	A notable difference in prediction quality appears between Monte Carlo sampling and the other methods.
	While there is still a high degree of overlap in the remaining methods, QMC-2 demonstrates a slight edge in the distribution of outcomes.\\
	}
	\postcaption
  \label{fig:resnet_quad}
\end{figure}

\Cref{fig:resnet_quad} shows the trajectories of prediction quality metrics for different quadratures, holding every other hyperparameter the same.
Monte Carlo (MC), quasi-Monte Carlo with first- and second-moment matching (QMC-1 and QMC-2, respectively), and the Hadamard cross-polytope sequence (Hadamard-Cross)
are compared with the same set of 8 random seeds.
The shaded region contains the full range of outcomes and the dotted line is the median.

The most significant training improvement is due to using quadratures that exactly integrate linear basis functions,
which captures the dominant contribution to the average gradient.
While second order effects have a small bearing on the training loss, there is strong overlap in the validation accuracies.
For this architecture, QMC-2 samples show a slight edge for the best runs.

\begin{figure}[ht]
	\centering
	\includegraphics[width=0.95\textwidth]{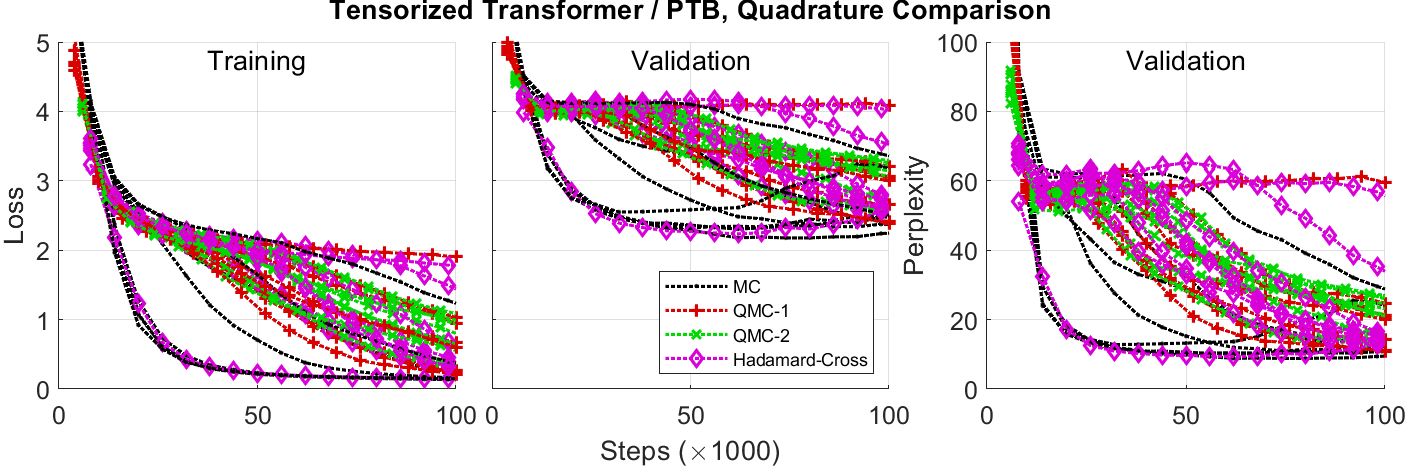}
	\precaption
	\caption{
	\textbf{Left:} Training loss. \textbf{Middle:} Validation loss. \textbf{Right:} Validation perplexity.
	Each training trajectory is shown because there is too much overlap in the range of outcomes
	for upper and lower bounds to be informative.
	While it is difficult to draw clear conclusions, several Monte Carlo runs exhibit the best early validation progress.
	}
	\postcaption
  \label{fig:ptb_quad}
\end{figure}

\paragraph{Tensorized Transformer / PTB}
All experiments use a three-layer tensorized transformer with a dropout rate of $0.3$,
$\beta_1 = 0.9$, $\beta_2 = 0.999$, and $\eps = 10^{-8}$.
Tuning for this architecture used 4 gradient evaluations with QMC-1 sampling.
The resulting hyperparameters are $\sigmaMin = 1\times10^{-5}$, $\sigmaMax = 4\times10^{-3}$, $\lambda=1\times10^{-3}$, and $w = 2\times10^6$.
No annealing or learning rate reduction schedules were tested.
From these hyperparameters, \Cref{fig:ptb_quad} shows 8 training runs for each quadrature using the same set of random seeds.
Several Monte Carlo runs achieve the best validation outcomes early on.
These tests demonstrate that none of the sampling or quadrature methods tested provide consistently superior results for all architectures.

\vspace{5mm}

\subsection{Algorithm Comparison}

\paragraph{ResNet18 / CIFAR-10}
These experiments compare stochastic gradient descent with momentum (SGD-M), Adam, AdaHessian, stochastic gradient variational Bayes (SGVB), and QNVB.
The SGD-M learning rate is $\lambda=0.1$ with a gradient momentum coefficient of $\beta_1=0.9$.
Adam runs use the standard hyperparameters, $\lambda=10^{-3}$, $\beta_1=0.9$, $\beta_2=0.999$, $\eps=10^{-8}$.
AdaHessian uses the standard learning rate of $\lambda=0.15$ and the remaining hyperparameters are the same as Adam.

\begin{figure}[ht]
	\centering
	\includegraphics[width=0.95\textwidth]{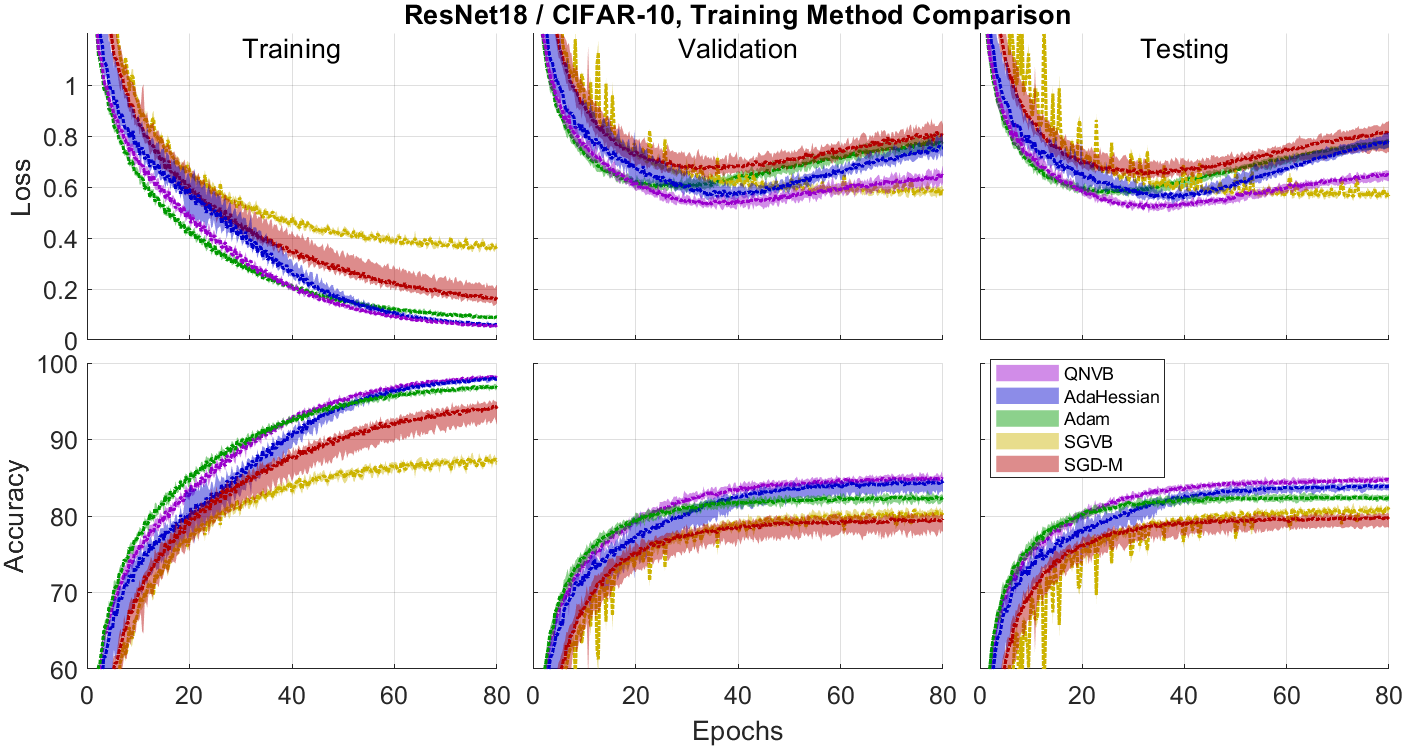}
	\precaption
	\caption{
	\textbf{Left column:} Training loss (top) and accuracy (bottom) trajectories.
	Both QNVB and AdaHessian achieve the lowest loss, and highest accuracy, for training data.\\
	\textbf{Middle column:} Validation loss and accuracy.
	QNVB achieves the lowest validation loss just before epoch 40.
	Accuracy continues to improve to the end of training and QNVB obtains consistently strong results.
	The best AdaHessian runs show some overlap with QNVB.\\
	\textbf{Right column:} Test loss and accuracy.
	This holdout partition demonstrates similar outcomes to the validation set.
	}
	\postcaption
  \label{fig:resnet}
\end{figure}

SGVB typically uses Monte Carlo samples to approximate the variational objective and then computes the gradient with respect to the variational parameters,
i.e.~the mean $\vMu$ and standard deviations $\vSigma$, which allows the variational parameters to be optimized with Adam.
Since SGVB is compatible with any quadrature or sampling method, the Hadamard cross-polytope sequence with 4 evaluations is used for both SGVB and QNVB to provide an equal comparison.
Further, the same lower and upper bounds on standard deviations ($\sigmaMin$ and $\sigmaMax$, respectively) are enforced with the tuned hyperparameters from \Cref{sub:quad}.
The learning schedule from \Cref{sub:quad} is also applied to all algorithms.
Results are shown in \Cref{fig:resnet}.

\paragraph{Tensorized Transformer / PTB}
These experiments test SGD-M, Adam, AdaHessian, SGVB, and QNVB using the same architecture and hyperparameters as in \Cref{sub:quad} with a few modifications that follow \cite{yao2021adahessian} and \cite{ma2019tensorized}.
The learning rates for SGD-M, Adam, and AdaHessian are $\lambda=5\times10^{-4}$, $\lambda=2.5\times10^{-4}$, and $\lambda=1$, respectively.
SGD-M uses a gradient momentum coefficient of $0.9$ and both SGD-M and Adam use a cosine learning rate schedule.
SGVB and QNVB use 4 gradient evaluations from the Hadamard cross-polytope sequence.

\begin{figure}[ht] 
	\centering
	\includegraphics[width=0.65\textwidth]{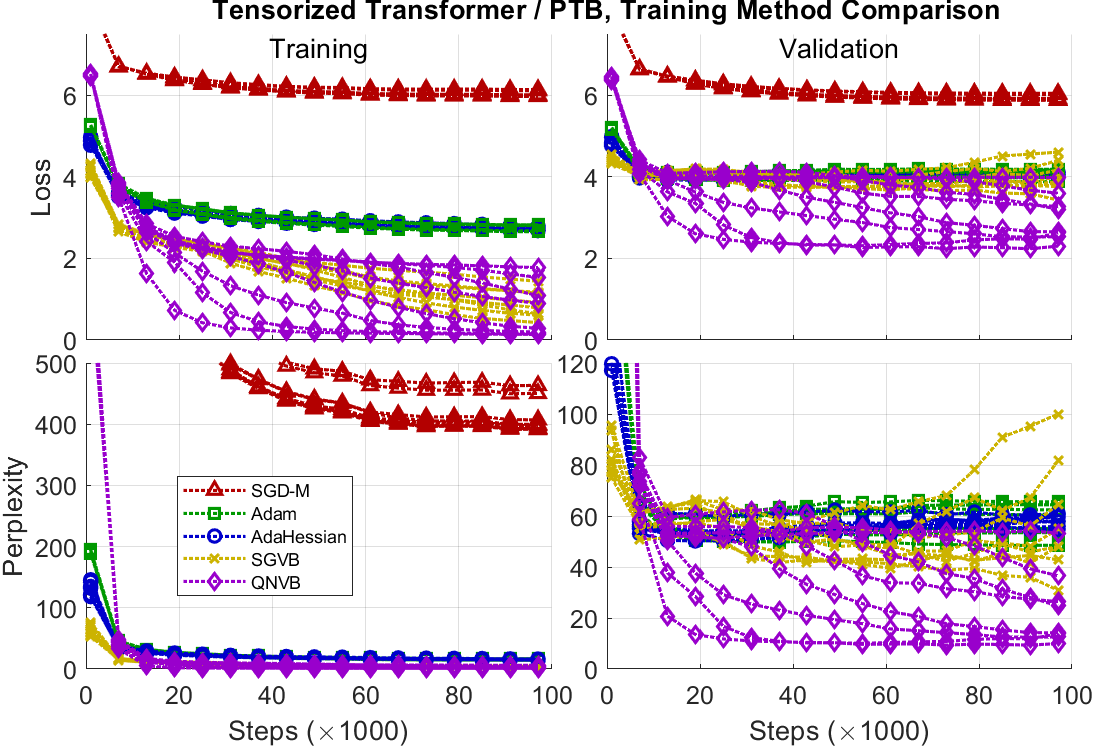}
	\precaption
	\caption{
	\textbf{Left column:} Training loss (top) and perplexity (bottom) trajectories.
	Since there is high variation in some outcomes, all 8 trajectories are shown for each algorithm.
	QNVB achieves the best outcomes followed by SGVB.
	Note, however, that some of the late training improvements shown by SGVB may be attributed to memorization, as they are not reflected by validation outcomes.\\
	\textbf{Right column:} Validation loss and perplexity.
	Since SGD-M is not competitive, it does not appear in this perplexity scaling.
	QNVB achieves state-of-the-art validation perplexity, with the best trial achieving 9.35.
	}
	\postcaption
  \label{fig:ptb}
\end{figure}

At the end of training, the optimal validation state is used to compute test perplexity.
Lower perplexity is better.
The optimal test perplexities are given in \Cref{fig:test_ppt}.

\begin{table}[ht]
\label{fig:test_ppt}
\centering
\caption{Tensorized Transformer / PTB, Optimal Test Perplexity}\vspace{-3mm}
\begin{tabular}{| r | r | r | r | r |} \cline{1-5}
SGD-M & Adam & AdaHessian & SGVB & QNVB \\ \cline{1-5}
343.1 & 42.12 & 45.18 & 29.02 & 8.89 \\ \cline{1-5}
\end{tabular}
\postcaption
\end{table}

\paragraph{Deep Learning Recommendation Model / Criteo Ad Kaggle}
These tests include stochastic gradient descend (SGD), AdaGrad \cite{duchi2011adaptive}, SGVB, and QNVB.
Both SGD and AdaGrad use the standard hyperparameters for DLRM.
AdaHessian could not be tested because this architecture includes sparse gradients during training, which do not provide the needed functionality for secondary backpropagation.
Yao et al.~\cite{yao2021adahessian} circumvent this issue by only applying AdaHessian to the parameters with dense gradients,
but that modification was not available for these experiments.

\begin{figure}[h!]
	\centering
	\includegraphics[width=0.95\textwidth]{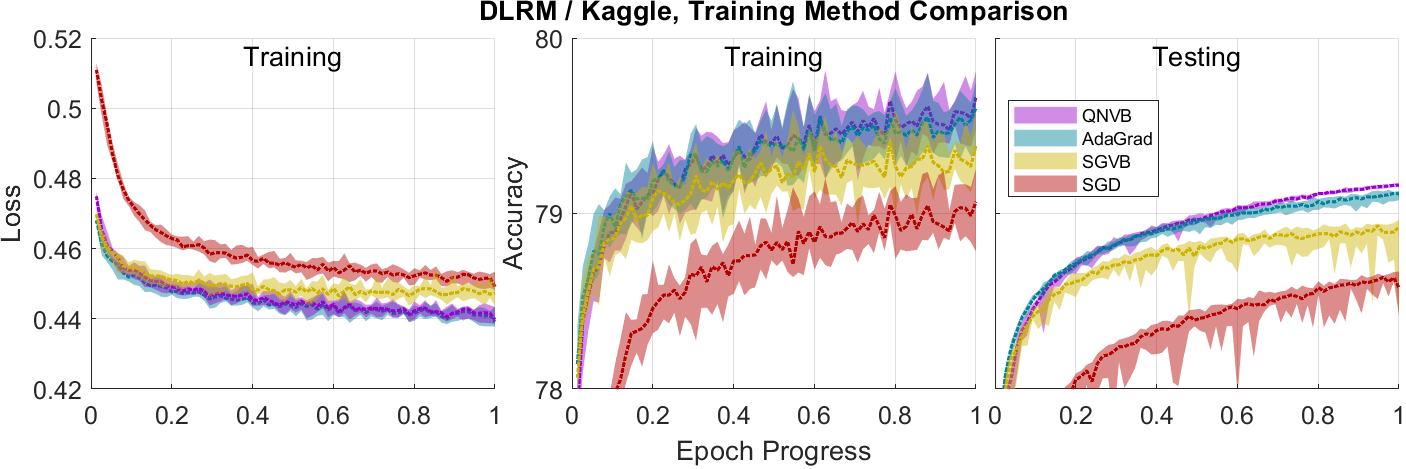}
	\precaption
	\caption{
	\textbf{Left}: Training loss. Both QNVB and AdaGrad achieve the lowest optimization outcomes for the training dataset.
	SGVB performs well initially, but this early trend slows.\\
	\textbf{Middle}: Training accuracy. There is significant overlap in optimal outcomes for both QNVB and AdaGrad.
	This is followed by SGVB and then SGD.\\
	\textbf{Right}: Testing accuracy. QNVB consistently achieves top accuracy scores on the test data.\\
	}
	\postcaption
  \label{fig:dlrm}
\end{figure}

Since DLRM is a very large model, with 540 million parameters, both SGVB and QNVB use 3 Monte Carlo samples per step to avoid excessive memory usage.
SGVB uses the standard Adam hyperparameters and $\sigmaMin=10^{-5}$ and $\sigmaMax=10^{-3}$.
QNVB uses the same lower and upper bounds as well as $\lambda=2\times10^{-3}$ and $w=5\times10^{-4}$.
QNVB also includes likelihood annealing, increasing the likelihood weight by a factor of $1.0000248$ per step to obtain a final weight of $w=10^8$.
It should also be noted that QNVB and SGVB require substantially more training time than SGD and AdaGrad.
This is because both SGD and AdaGrad take advantage of sparse gradients to avoid updating all parameters with each step.
In contrast, both SGVB and QNVB must modify each parameter multiple times per step, requiring approximately 13 hours to complete the full epoch of $300,000$ training batches.
Results are shown in \Cref{fig:dlrm}.

This architecture exhibits extreme susceptibility to memorization, with validation outcomes immediately deteriorating in a second epoch,
constraining optimization to only allow a single pass over the training data.
QNVB obtains the best testing accuracy, with a top result of $79.173\%$,
which is marginally better than the reported result of $79.167\%$ for AdaHessian.

\subsection{Discussion}

\paragraph{Hessian approximations}
Although the quadrature can make a difference in prediction quality for certain architectures, there does not appear to be a consistently superior scheme.
This is likely due to the fact that approximations errors are not only due to the evaluation locations, but also due to different training samples constraining different parameters.
For the average Hessian to account for these topography constraints, which may only appear intermittently, the running average must use many effective sample batches.
Further, by tracking the Hessian squared, the average is dominated by the largest-magnitude samples in the average,
reducing the impact of the error averaging properties.
As a result, it is recommended to simply test feasible variations during tuning.

\paragraph{Mean-field densities}
Mean fields are somewhat simplistic approximations of the posterior.
Viewed from the context of tensors, mean-field distributions are equivalent to rank-1 functions \cite{Trefethen2017}.
We can compare a rank-$r$ Canonical Polyadic (CP) approximation, a rank-$r$ tensor, to a mean-field mixture as 
\begin{align*}
\mathcal{X}_j &\approx \sum_{k=1}^r \lambda_k \prod_{m=0}^{d-1} A^{(m)}_{j_m k} &
\pP(\vPa) &\approx \sum_{k=1}^r \pQ(k) \prod_{i=0}^{d-1} \pQ( \vPa_i \mid k ).
\end{align*}
Although this work does not explore higher-rank variational distributions, the resulting mean-field mixtures would also be scalable and feasible to integrate,
only multiplying the integration cost by the rank, i.e.~the number of mixture components.

\paragraph{Other Bases}
Although the analysis in \Cref{sec:variational} only examines Gaussian mean-field distributions and Dirac-Gauss mixtures,
a related approach may be suitable for Laplacian mean-field densities or perhaps even Dirac-Laplace mixtures.
One difficulty, however, is that the quadratic basis functions, $(\vPa - \vNu)^2$, would have to be replaced by absolute value functions, $\left| \vPa - \vNu \right|$.
Unfortunately, this would require an adaptive basis, which may present additional analytic challenges.

\subsection{Summary}

This work began by investigating stochastic blocked mean-field quadratures in order to ensure numerical integration would retain high accuracy within specific parameter blocks,
while also ensuring many samples converge to a tensor-product cubature.
By considering quasirandom sequences to speed up convergence, we arrived at quadrature sequences composed of antithetic evaluation pairs from the cross-polytope sequence in the Hadamard basis.
In $d$ dimensions, this method exactly integrates approximately $\frac{1}{4}d^2$ multivariate quadratic basis functions using only 4 function evaluations,
which gives the highest exactness efficiency for all the methods tested.
\Cref{thm:exactness_periodicity} shows how every doubling of the number of evaluations increases exactness to include half of the remaining quadratic basis functions.

In order to use these quadratures efficiently, we examined the optimal structure of variational densities that can be written as an exponential of a linear combination of basis functions,
which yielded projective integral updates and a corresponding fixed-point formulation of the optimizer.
Adjusting the quadrature sequences to act on gradients allows us to approximate the needed projection integrals, comprising average gradients and Hessian diagonals,
for Gaussian mean-fields.
Related analysis for Dirac-Gauss mixtures may prove useful for sparsifying variational inference.
We then introduce algorithmic details for quasi-Newton variational Bayes, a practical implementation of projective integral updates for Gaussian mean fields that can be used to train deep learning architectures.

Experiments show that training with different quadratures can have a significant impact on outcomes.
QNVB also demonstrates the ability to compete with state-of-the-art training algorithms on three different learning problems and architectures.
This work provides new approaches to calibrate and control model uncertainty during training,
which can improve generalizability by capturing plausible model variations and better navigate the loss topography during optimization.
 
\subsection*{Acknowledgements}
My sincere thanks to Erin Acquesta, Tommie Catanach, Jaideep Ray, Cosmin Safta, and Jacquilyn Weeks for providing early feedback.
Tommie suggested an exponential annealing schedule and log-likelihood integration experiments.
Jaideep noted that by limiting the diameter of evaluation nodes,
these quadrature sequences may provide an additional benefit to prediction models that cannot support large parameter perturbations,
such as physics models that cannot accept unphysical states.
Special thanks go to Alexander Safonov for consultation on the PyTorch implementation of QNVB and for helping curate the Tensorized Transformer test scripts.

Sandia National Laboratories is a multi-mission laboratory managed and operated by National Technology \& Engineering Solutions of Sandia, LLC (NTESS), a wholly owned subsidiary of Honeywell International Inc., for the U.S. Department of Energy’s National Nuclear Security Administration (DOE/NNSA) under contract DE-NA0003525. This written work is authored by an employee of NTESS. The employee, not NTESS, owns the right, title and interest in and to the written work and is responsible for its contents. Any subjective views or opinions that might be expressed in the written work do not necessarily represent the views of the U.S. Government. The publisher acknowledges that the U.S. Government retains a non-exclusive, paid-up, irrevocable, world-wide license to publish or reproduce the published form of this written work or allow others to do so, for U.S. Government purposes. The DOE will provide public access to results of federally sponsored research in accordance with the DOE Public Access Plan.

\appendix

\section{Numerical Integration Experiments}
\label{sec:numerical_integration}

\begin{algorithm}[b!]
\caption{Simplex Polytope Sigma Points}
\label{alg:simplex}
\fontsize{10}{16}\selectfont
\begin{algorithmic}[1]
\Require $\nDim$ is the number of dimensions in which the desired sigma points are embedded.
\Ensure $\mX$ is $\nDim \times \nDim+1$ matrix of evaluation nodes and corresponding weights, $w$.
\Function{$(\mX,w) = \fSimplexQuad$}{$\nDim$}
\State $r = \sqrt{\nDim}$
\State $\mX = 0^{\nDim \times \nDim+1}$
\For{$\iDim = 1, 2, \ldots, \nDim$}
   \State $\mX_{\iDim \iDim} = r$ and $\mX_{\iDim j} = \frac{-r}{\nDim+1-\iDim}$ for $j=\iDim+1, \iDim+2, \ldots, \nDim+1$
   \State $r \gets r \frac{\sqrt{(\nDim+1-\iDim)^2 - 1}}{\nDim+1-\iDim}$
\EndFor
\State Return $\mX$ and shared weight, $w = \frac{1}{\nDim+1}$.
\EndFunction
\end{algorithmic}
\end{algorithm}

\begin{figure}[t!]
	\centering
	Quadrature Error for Mean-Field Gaussian 
	\includegraphics[width=0.90\textwidth]{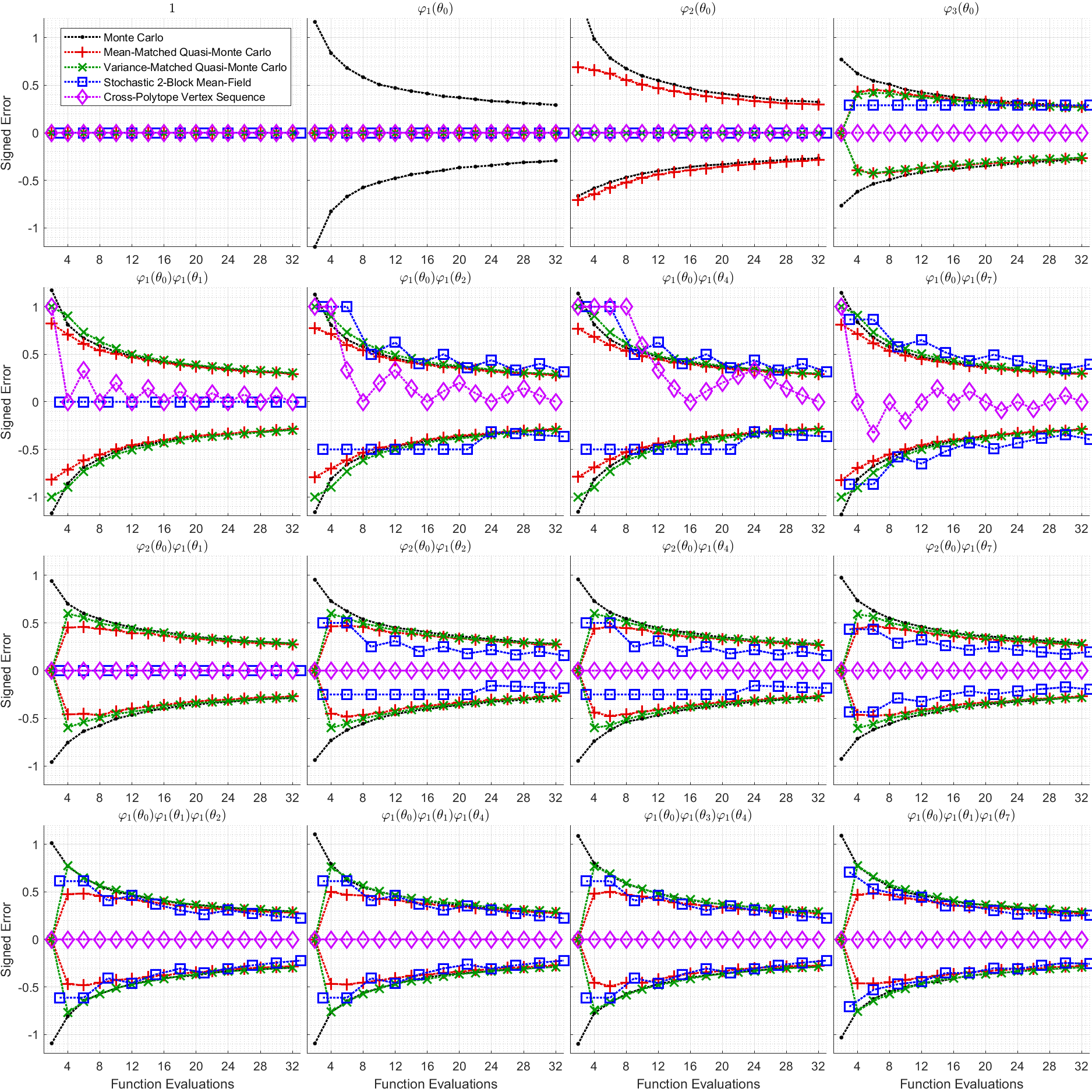}
	\precaption
	\caption{
	Selected quadrature errors for, $\pQ(\vPa_i) \equiv \prod_{\iPa=0}^\nPa \pN(\vPa_i \mid 0, 1)$.
	Each plot shows the quadrature error for products of orthonormal polynomials, $\varphi_d(\cdot)$, where $d$ indicates the degree.\\
	\textbf{Row 1}: Univariate basis functions show how quasi-Monte Carlo methods achieve exactness on first and second degree polynomials.
	These methods do not necessarily reduce error on higher-order basis functions; compare to \Cref{fig:dirac_gauss}.
	The stochastic 2-block mean-field quadrature does not correctly integrate 3\textsuperscript{rd}-order basis functions.
	The cross-polytope sequence is exact for these functions.\\
	\textbf{Row 2}: Quasi-Monte Carlo methods offer little improvement to these multivariate quadratic integrals.
	Stochastic blocking retains exactness for quadratics within each block, column 1, but not between blocks.
	Finally, the cross-polytope sequence shows the exactness periodicity we expect.\\
	\textbf{Row 3}: Stochastic blocking appears to reduces the error for these 3\textsuperscript{rd}-order functions.
	The cross-polytope sequence produces Gauss points, roots of $\varphi_2(\vPa_0)$, causing all products to vanish.\\
	\textbf{Row 4}: The cross-polytope sequence generates odd pairs of evaluations for each odd basis function.
	The product of an odd number of such evaluations remains odd, thus correctly summing to zero.
	This is why both quasi-Monte Carlo methods integrate to zero with two samples.
	}
	\postcaption
  \label{fig:gauss}
\end{figure}

This set of experiments, \Cref{fig:gauss,fig:laplace,fig:dirac_gauss}, shows the typical range of integration errors for several numerical approaches.
The first approach is pure Monte Carlo integration via sampling the mean-field distribution.
The second and third approaches are quasi-Monte Carlo, translating the set of samples to match the mean (second) and also scaling to match both the mean and the variance (third).

The fourth approach demonstrates stochastic blocked mean-field quadratures with a block size of 2.
This requires 3 sigma points, given by the simplex vertices from \Cref{alg:simplex}, within in each 2D block.
Within each block, the evaluation nodes are then permuted uniformly at random and concatenated.

The fifth approach shows antithetic pairs of cross-polytope vertices in the Hadamard basis, \Cref{alg:cross-polytope}. 
Signed errors are stored and sorted for each integration method from 5000 trials to obtain the 90\% confidence intervals.

\subsection{Symmetric Distributions}

\begin{figure}[t!]
	\centering
	Quadrature Error for Mean-Field Laplace 
	\includegraphics[width=0.90\textwidth]{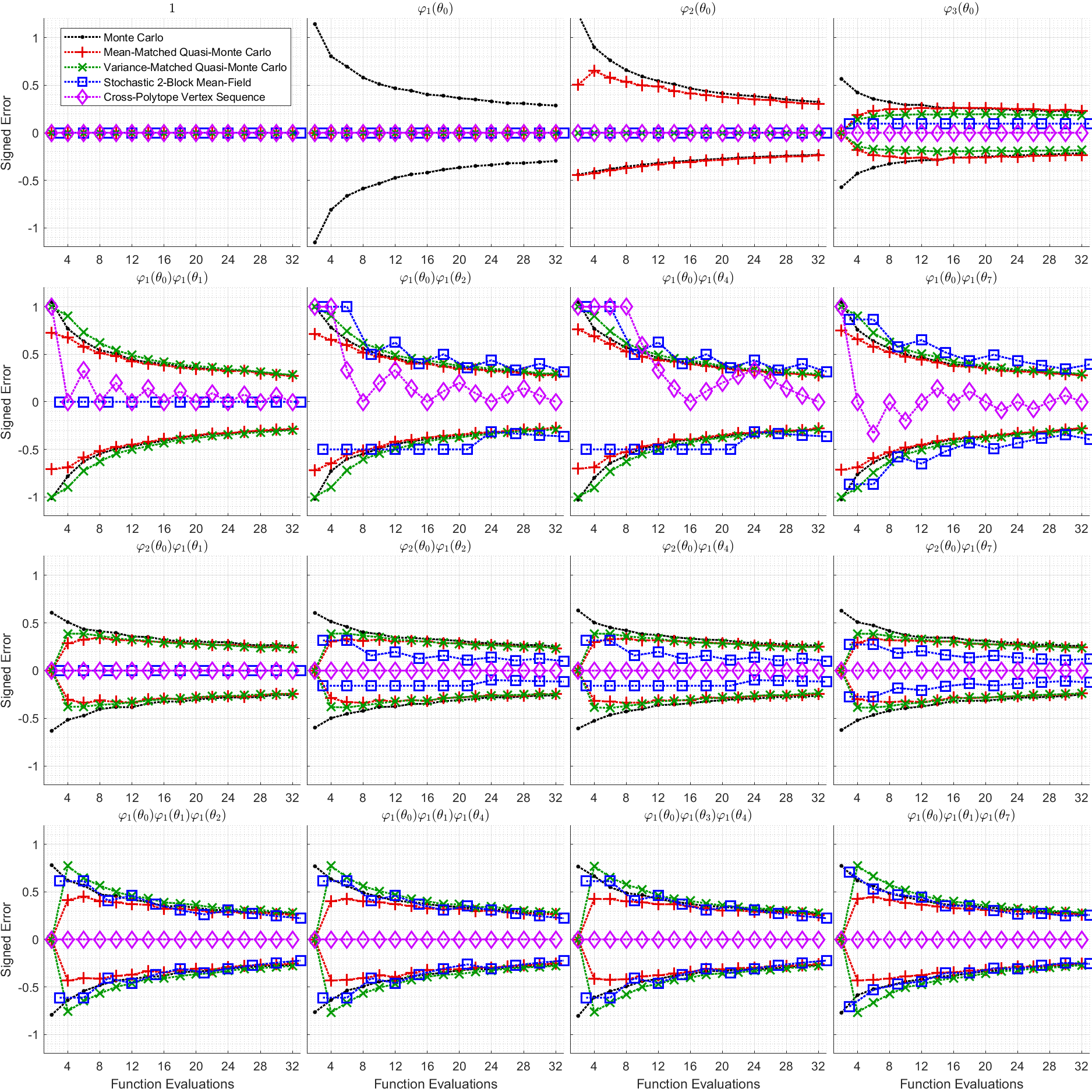}
	\precaption
	\caption{
	Selected quadrature errors for, $\pQ(\vPa_i) \equiv \prod_{\iPa=0}^\nPa \pL(\vPa_i \mid 0, 1)$.
	The critical property of the mean-field distribution that determines the characteristics of these error plots is symmetry.
	Since Laplace distributions are symmetric, we observe the same structures as the Gaussian case.
	The only notable difference is the scale of errors for some of the stochastic methods.
	For example, errors associated with $\varphi_3(\vPa_0)$ (top-right) are significantly smaller in this case,
	as are the errors for the multivariate cubics in row 3.
	The cross-polytope sequence quadratures are exact for all the same cases as before.
	}
	\postcaption
  \label{fig:laplace}
\end{figure}

The first two mean-field distributions are Gaussian and Laplacian, \Cref{fig:gauss} and \Cref{fig:laplace}, respectively.
Both are both symmetric and demonstrate higher-order exactness for the cross-polytope quadrature sequences.

The univariate basis functions in Row 1 show how quasi-Monte Carlo methods achieve exactness on first and second degree polynomials
by transforming samples to match leading moments.
These methods may also reduce error on higher-order basis functions, but not always.
See \Cref{fig:dirac_gauss}.
The stochastic blocked mean-field quadrature does not correctly integrate 3\textsuperscript{rd}-order basis functions.
In contrast, the cross-polytope sequence in the Hadamard basis, always produces a pair of Gauss-points in each dimension,
thus integrating all univariate 3\textsuperscript{rd}-order polynomials exactly.

The multivariate quadratics in Row 2 show that even variance-matched quasi-Monte Carlo may not significantly improve mixed second-order integrals.
Stochastic blocking retains exactness for quadratics within each block, column 1, but not between blocks.
Finally, the cross-polytope sequence shows the exactness periodicity we expect, based on the leading mismatched bit between each pair of parameter indices.
For example, $\varphi_1(\vPa_0)\varphi_1(\vPa_7)$ has the same exactness periodicity as $\varphi_1(\vPa_0)\varphi_1(\vPa_1)$ because the bit strings, $000$ and $111$, differ in the leading bit,
just as $000$ and $001$.

In Row 3, the stochastic mean-field quadrature reduces error for these 3\textsuperscript{rd}-order functions.
Since the cross-polytope sequence produces Gauss points in each dimension, the zeros of $\varphi_2(\vPa_0)$,
all of these products correctly vanish.

The cross-polytope sequence performs well in Row 4 because it generates odd evaluation pairs for each odd function.
Since the product of an odd number of such evaluations is still odd, the average sums to zero.
This also explains why both moment-matching methods integrate to zero when they only contain two samples.

\subsection{Dirac-Gauss Mixture}

\begin{figure}[t!]
	\centering
	Quadrature Error for Mean-Field Dirac-Gauss Mixture 
	\includegraphics[width=0.90\textwidth]{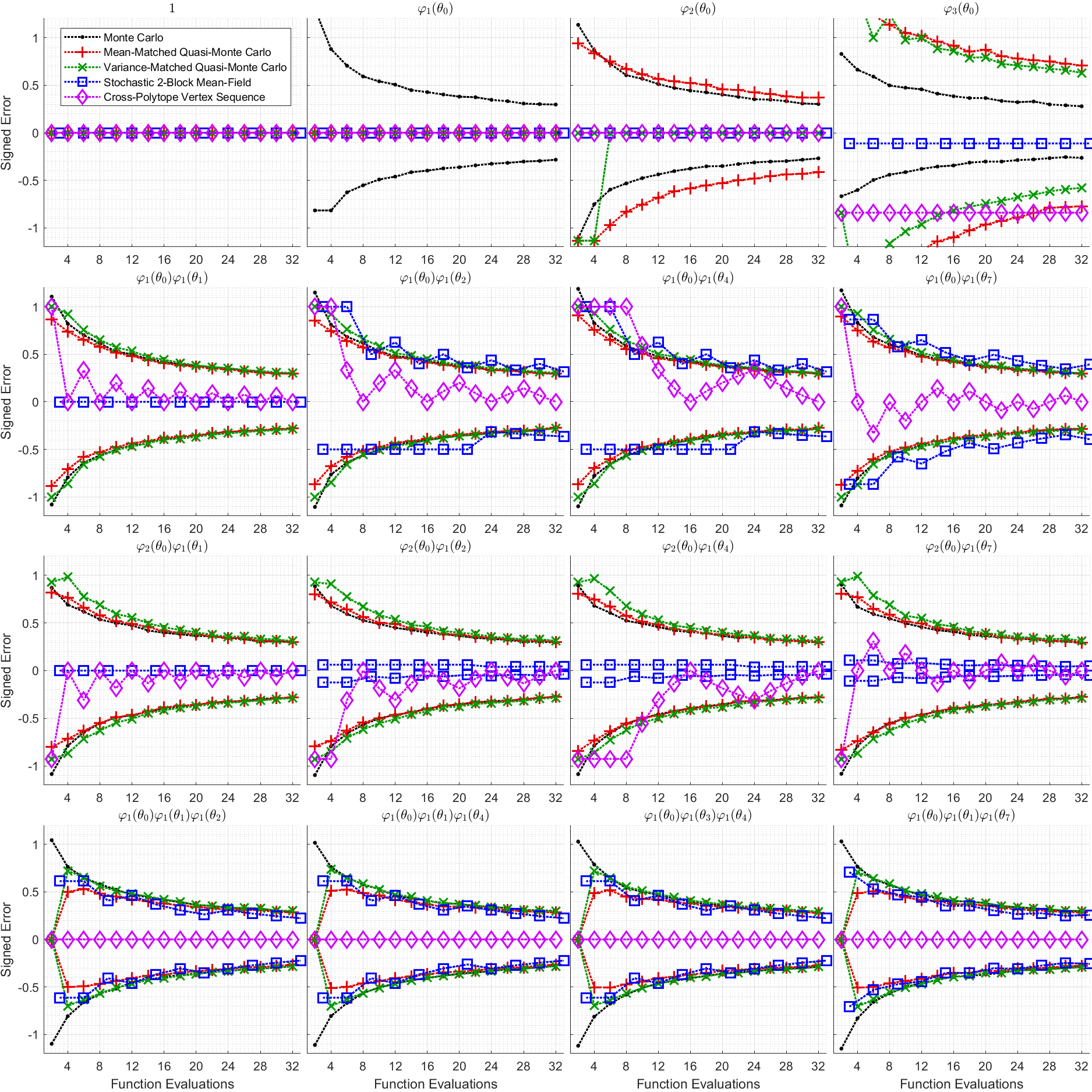}
	\precaption
	\caption{
	Selected quadrature errors for, $\pQ(\vPa_i) \equiv \prod_{\iPa=0}^\nPa \left( \frac{1}{2} \pD(\vPa_i) + \frac{1}{2}\pN(\vPa_i \mid 2, 1)\right)$.\\
	\textbf{Row 1}: Note the conspicuous errors for the few-sample variance-matched quasi-Monte Carlo cases with $\varphi_2(\vPa_0)$.
	Since the cross-polytope sequence does not produce Gaussian quadratures in each coordinate, we no longer obtain 3\textsuperscript{rd}-order exactness.\\
	\textbf{Row 2}: The cross-polytope sequence still operates as designed to integrate multivariate quadratics with the same periodicity as the previous cases.\\
	\textbf{Row 3}: The cross-polytope sequence is no longer always exact for these cases because it no longer evaluates at roots of $\varphi_2(\vPa_i)$,
	but we still obtain the same exactness periodicity as Row 2.\\
	\textbf{Row 4}: The cross-polytope sequence is still exact for these cases for the same reason as before. 
	}
	\postcaption
  \label{fig:dirac_gauss}
\end{figure}

The third mean-field distribution tested is a spike and slab, \Cref{fig:dirac_gauss}, which is not symmetric.
As a consequence of the asymmetry, it is not possible for the cross-polytope sequence to generate Gaussian quadrature pairs that also have equal-weights,
the essential property that allowed the cross-polytope sequence to generate 3\textsuperscript{rd}-order cubatures earlier.
We can still construct equal-weight quadratures for this purpose, but they are only 2\textsuperscript{nd}-order in each dimension,
thus only becoming second-order cubatures with the exactness periodicity of \Cref{thm:exactness_periodicity}.

The errors in Row 1 for the few-sample variance-matched quasi-Monte Carlo cases with $\varphi_2(\vPa_0)$
occur because each factor distribution contains finite probability mass at $\vPa_i=0$.
With only a few samples, a specific coordinate is often zero for all samples, meaning it is not possible to match the sample variance to the distribution variance.
Only being able to match the mean causes these results.
Again, since the cross-polytope sequence does not produce Gaussian quadratures in each coordinate, we no longer obtain cubic exactness.

Row 2 shows that the cross-polytope sequence still operates as designed to integrate multivariate quadratics with the same periodicity as the previous cases.
Row 3, however, shows that the cross-polytope sequence is no longer always exact for these cases, since it no longer evaluates roots of $\varphi_2(\vPa_i)$.
Instead, we revert to the same exactness periodicity as seen in Row 2.
In Row 4, the same reasoning for odd products of odd function evaluations still holds, so these are still exact.

\section{Proofs}
\label{sec:proofs}

\subsection{Proof of Theorem \ref{thm:expect}}
\label{sub:pf_expect}

Since all permutations are equally likely, we have
\begin{align*}
& \expect_{P_1, P_2, \ldots, P_\nBlock} f(\vPa^{(1)}) = \frac{1}{\nQuad^\nBlock} \sum_{\iQuad_1=1}^\nQuad \sum_{\iQuad_2=1}^\nQuad \cdots \sum_{\iQuad_\nBlock=1}^\nQuad \prod_{\iBlock=1}^\nBlock f^{(\iBlock)}(\vPa_\iBlock^{(\iQuad_\iBlock)}) \\
&= \left( \frac{1}{\nQuad} \sum_{\iQuad_1=1}^\nQuad f^{(1)}(\vPa_1^{(\iQuad_1)}) \right) \left( \frac{1}{\nQuad} \sum_{\iQuad_2=1}^\nQuad f^{(2)}(\vPa_2^{(\iQuad_2)}) \right) \cdots \left( \frac{1}{\nQuad} \sum_{\iQuad_\nBlock=1}^\nQuad f^{(\nBlock)}(\vPa_\nBlock^{(\iQuad_\nBlock)}) \right) \\
&= \left( \int f^{(1)}(\vPa_1)\, d\pQ(\vPa_1) \right) \left( \int f^{(2)}(\vPa_2)\, d\pQ(\vPa_2) \right) \cdots \left( \int f^{(\nBlock)}(\vPa_\nBlock)\, d\pQ(\vPa_\nBlock) \right) \\
&= \int f(\vPa)\, d\pQ(\vPa).
\end{align*}
Since this holds for all evaluation nodes, the average yields the same result${}_\square$

\subsection{Proof of Lemma \ref{lem:relative_parity}}
\label{sub:pf_relative_parity}

\begin{align*}
& \vP_{\iPa_1} \xor \vP_{\iPa_2} = 
\left( \xor_{\iBit=1}^{\nBit} \left[ \bit_{\iBit}(\iPa_1) \and \bit_{\iBit}(\iQuad) \right] \right) \xor \left( \xor_{\iBit=1}^{\nBit} \left[ \bit_{\iBit}(\iPa_2) \and \bit_{\iBit}(\iQuad) \right] \right) \\
&= \xor_{\iBit=1}^{\nBit} \left[ \left( \bit_{\iBit}(\iPa_1) \and \bit_{\iBit}(\iQuad) \right) \xor \left( \bit_{\iBit}(\iPa_2) \and \bit_{\iBit}(\iQuad) \right)\right] \\
&= \xor_{\iBit=1}^{\nBit} \left[ \left( \bit_{\iBit}(\iPa_1) \xor \bit_{\iBit}(\iPa_2) \right) \and \bit_{\iBit}(\iQuad) \right] = \xor_{\iBit=1}^{\nBit} \left[ \vX_{\iBit} \and \bit_{\iBit}(\iQuad) \right]_\square
\end{align*}

\subsection{Proof of Theorem \ref{thm:exactness_periodicity}}
\label{sub:pf_exactness_periodicity}

This result easily follows from \Cref{lem:relative_parity}.
As $\iQuad$ increases, the relative parity, $\vP_{\iPa_1} \xor \vP_{\iPa_2}$, can only switch when at least one bit, $\bit_{\iBit}(\iQuad)$, at a position $\iBit \geq b$ flips.
Thus, starting at $\iQuad = z 2^b$,
we must obtain $2^{b-1}$ iterates of the same parity followed by another $2^{b-1}$ iterates of the other parity.
Thus, when we sum corresponding evaluation nodes for any function $f(\vTh)$ that only depends on $\vPa_{\iPa_1}$ and $\vPa_{\iPa_2}$,
we must obtain equal-weight contributions from both antithetic pairs,
\begin{align*}
& \frac{1}{2^{b+1}} \sum_{\iQuad = z 2^b}^{(z+1) 2^b - 1} f(\vPa^{(2\iQuad+1)}) + f(\vPa^{(2\iQuad+2)}) \\
& = \frac{1}{4}\left( f(\vPa_{\iPa_1} = \vMu_{\iPa_1} + \vSigma_{\iPa_1}, \vPa_{\iPa_2} = \vMu_{\iPa_1} + \vSigma_{\iPa_2} )
+ f(\vPa_{\iPa_1} = \vMu_{\iPa_1} - \vSigma_{\iPa_1}, \vPa_{\iPa_2} = \vMu_{\iPa_1} - \vSigma_{\iPa_2} ) \right. \\
&  \quad\,\,\,\left. + f(\vPa_{\iPa_1} = \vMu_{\iPa_1} + \vSigma_{\iPa_1}, \vPa_{\iPa_2} = \vMu_{\iPa_1} - \vSigma_{\iPa_2} )
+ f(\vPa_{\iPa_1} = \vMu_{\iPa_1} - \vSigma_{\iPa_1}, \vPa_{\iPa_2} = \vMu_{\iPa_1} + \vSigma_{\iPa_2} ) \right)_\square
\end{align*}

\subsection{Proof of Lemma \ref{lem:norm_preserving}}
\label{sub:pf_norm_preserving}

The proof easily follows by taking the Gateaux derviative with respect to the normalization,
\begin{align*}
&\frac{\partial}{\partial \eps} \left[ \int d\pQ(\vPa \mid \vPhi = \vPhi^* + \eps \vEta) \right]_{\eps=0}
= \int \sum_{\iBasis=0}^\nBasis \vEta_\iBasis \fBasis_\iBasis(\vPa)\, d\pQ(\vPa \mid \vPhi^*) \\
&= \sum_{\iBasis=0}^\nBasis \vEta_\iBasis \langle \fBasis_0, \fBasis_\iBasis \rangle_{\vPhi^*} = \vEta_0 = 0.
\end{align*}
Thus, by using an orthogonal basis and capturing the normalization coefficient with $\fBasis_0$, normalization-preserving perturbation directions only require fixing $\vEta_0=0_\square$

\subsection{Proof of Theorem \ref{thm:projective_integral}}
\label{sub:projective_integral}
Provided both the variational optimizer and the posterior distribution have full support over the parameter domain,
we can rewrite the posterior distribution by factoring out the optimizer and defining what remains as the residual, $\fResidual(\vPa)$, so that
\begin{align*}
\pP(\vPa \mid \sData) = \exp\left[\fResidual(\vPa) + \sum_{\iBasis=0}^n \vPhi_\iBasis^* \fBasis_\iBasis(\vPa)\right].
\end{align*}
Since arbitrary perturbations must satisfy the variational principle, we have
\begin{align*}
&\frac{\partial}{\partial \eps} \left[ \int \log\!\left( \frac{\pQ(\vPa \mid \vPhi^* + \eps \vEta)}{\pP(\vPa \mid \sData)} \right)\, d\pQ(\vPa \mid \vPhi^* + \eps \vEta) \right]_{\eps=0} \\
&= \int \left[ \sum_{\iBasis=1}^\nBasis \vEta_\iBasis \fBasis_\iBasis(\vPa) \right] \left[ 1 - \fResidual(\vPa) \right]\, d\pQ(\vPa \mid \vPhi^*) 
= -\sum_{\iBasis=1}^\nBasis \vEta_\iBasis \langle \fBasis_\iBasis, \fResidual(\vPa) \rangle_{\vPhi^*} = 0.
\end{align*}
Note that we have dropped $\vEta_0$ terms by applying \Cref{lem:norm_preserving}.
As each remaining $\vEta_\iBasis$ is arbitrary, all inner products must vanish. 
It follows that the residual must be orthogonal to the span of the variational basis, excluding the normalizing component in $\fBasis_0$,
which is typically unknown.

\bibliographystyle{./siamplain}
\bibliography{./refs_09_07_23}
\end{document}